\PassOptionsToPackage{numbers,sort&compress}{natbib}
\documentclass[sn-mathphys]{sn-jnl}
\usepackage{natbib}
\usepackage{graphicx}%
\usepackage{longtable,tabularx}
\usepackage{amsmath,amssymb,amsfonts}%
\usepackage{amsthm}%
\usepackage{mathrsfs}%
\usepackage[title]{appendix}%
\usepackage{xcolor}%
\usepackage{textcomp}%
\usepackage{adjustbox}
\usepackage{manyfoot}%
\usepackage{booktabs}\let\cline\cmidrule
\usepackage{amsmath}
\usepackage{algorithm}%
\usepackage{algorithmicx}%
\usepackage{algpseudocode}%
\usepackage{arydshln}
\usepackage{listings}%
\usepackage{nomencl}
\makenomenclature

\usepackage{subfig}
\usepackage{graphicx}
\usepackage{amsmath}
\usepackage[version=4]{mhchem}
\usepackage{siunitx}
\usepackage{longtable,tabularx}
\usepackage{hyperref}
\usepackage{dsfont}

\newcommand{\eqnref}[1]{Eq.~(\ref{#1})}
\newcommand{\figref}[1]{Fig.~\ref{#1}}
\newcommand{\tabref}[1]{Table~\ref{#1}}
\usepackage [autostyle,english = american]{csquotes}
\MakeOuterQuote{"}
\usepackage{tikz}
\usepackage{forest}
\usepackage{stanli}
\usetikzlibrary{positioning, shapes, shadows, arrows}

\newtheorem{theorem}{Theorem}

\usepackage{hyperref}
\hypersetup{
    unicode = false,
    pdftoolbar = true,
    pdfmenubar = true,
    pdffitwindow = true,
    pdftitle = {meta},
    pdfauthor = {Saves et al.},
    pdfsubject = {meta},
    pdfkeywords = {meta},
    pdfnewwindow = true,
    colorlinks = true,
    linkcolor = blue,
    citecolor = blue,
    filecolor = black,
    urlcolor = blue,
    breaklinks = true
}

\usepackage[utf8]{inputenc}
\usepackage{comment}
\usepackage[english]{babel}
\usepackage[T1]{fontenc}
\usepackage[dvipsnames]{xcolor}
\usepackage{lmodern}
\usepackage{anyfontsize}
\usepackage{forest}
\usepackage[normalem]{ulem}
\usepackage{ragged2e}
\usepackage{amsmath}

\usepackage{stackengine}
\usepackage{mathtools}
\usepackage{stackrel}
\usepackage{dsfont}

\usepackage{mathrsfs}   
\usepackage{changepage}   
\usepackage{enumitem}

\usepackage{amsfonts}
\usepackage{hyperref}
\usepackage{makecell}
\usepackage{amssymb}
\usepackage{lipsum}
\usepackage{multicol}
\usepackage[font=small]{caption}
\usepackage[separate-uncertainty=true]{siunitx}
\usepackage{amsmath}
\usepackage{graphicx}
\usepackage{tikz}
\usepackage{forest}
\usepackage{stanli}
\usetikzlibrary{positioning, shapes, shadows, arrows, automata, arrows.meta, decorations, decorations.text, calligraphy}
\usepackage{pgfplots}
\pgfplotsset{compat=1.18}

\usepackage{fancyhdr}

\newcommand{\tmpx}{}

\fancypagestyle{sec}{\lhead{\tmpx}}

\def\getangle(#1)(#2)#3{%
  \begingroup%
    \pgftransformreset%
    \pgfmathanglebetweenpoints{\pgfpointanchor{#1}{center}}{\pgfpointanchor{#2}{center}}%
    \expandafter\xdef\csname angle#3\endcsname{\pgfmathresult}%
  \endgroup%
}

\usepackage{xfrac}
\usepackage{adjustbox}
\usepackage{standalone}
\usepackage{changepage}
\usepackage{yhmath}

\usepackage[textwidth=20mm,backgroundcolor=yellow,linecolor=orange,textsize=tiny]{todonotes}

\providecommand{\orcidlink}[1]{%
  \href{https://orcid.org/#1}{\scalebox{0.05}{\orcidlogo}}%
}

\definecolor{Red}{rgb}{1,0,0}
\definecolor{Green}{rgb}{0,.6,0}
\definecolor{black}{rgb}{0,0,0}

\newcommand{\meta}{\mathrm{m}}        
\newcommand{\metadec}{\mathrm{md}}    
\newcommand{\neutral}{ \mathrm{neu} }
\newcommand{\decreed}{\mathrm{dec}}

\newcommand{\quant}{\mathrm{qnt}}
 
\newcommand{\cat}{\mathrm{cat}}

\newcommand{\dist}{\mathrm{dist}}
\newcommand{\edist}{\overline{\mathrm{dist}}}

\newcommand{\acting}{\mathrm{inc}}

\renewcommand{\thesection}{\arabic{section}}

\usepackage{amsthm}
\theoremstyle{plain}
\newtheorem{mydef}{Definition}

\usepackage{listings}
\hypersetup{
    colorlinks,
    linkcolor={red!20!black},
    citecolor={blue!20!black},
    urlcolor={blue!80!black}
}

\begin{document}

\newcommand{\Title}
{\textcolor{black}{Modeling Hierarchical Spaces: A Review and Unified Framework for Surrogate-Based Architecture Design}}
\title[\Title]{\Title}

\author*[1]{\fnm{Paul} \sur{Saves}\orcidlink{0000-0001-5889-2302} }\email{paul.saves@irit.fr}
\author[2]{Edward Hallé-Hannan\orcidlink{0009-0007-7354-8746}}
\author[3]{Jasper Bussemaker\orcidlink{0000-0002-5421-6419}}
\author[2]{Youssef Diouane\orcidlink{0000-0002-6609-7330}}
\author[4,5]{Nathalie Bartoli\orcidlink{0000-0002-6451-2203}}

\affil[1]{IRIT and UT Capitole, Université de Toulouse, CNRS, Toulouse INP, UT3, UT2J, UT Capitole, Toulouse, France}
\affil[2]{GERAD and Department of Mathematics and Industrial Engineering, Polytechnique Montréal, Montréal, QC, Canada}
\affil[3]{Institute of System Architectures in Aeronautics, German Aerospace Center (DLR), Hamburg, Germany}
\affil[4]{DTIS, ONERA, Université de Toulouse, 31000, Toulouse, France}
\affil[5]{Fédération ENAC ISAE-SUPAERO ONERA, Université de Toulouse, 31000, Toulouse, France}

\email{edward.halle-hannan@polymtl.ca, jasper.bussemaker@dlr.de,  youssef.diouane@polymtl.ca, nathalie.bartoli@onera.fr}

\abstract{
Simulation-based problems involving mixed-variable inputs frequently feature domains that are hierarchical, conditional, heterogeneous, or tree-structured. These characteristics pose challenges for data representation, modeling, and optimization. This paper reviews extensive literature on these structured input spaces and proposes a unified framework that generalizes existing approaches.
In this framework, input variables may be continuous, integer, or categorical. A variable is described as \emph{meta} if its value governs the presence of other \emph{decreed variables}, enabling the modeling of conditional and hierarchical structures. We further introduce the concept of \emph{partially-decreed variables}, whose activation depends on contextual conditions.
To capture these inter-variable hierarchical relationships, we introduce \emph{design space graphs}, combining principles from feature modeling and graph theory. This allows the definition of general hierarchical domains suitable for describing complex system architectures. 
\textcolor{black}{Our framework defines hierarchical distances and kernels to enable surrogate modeling and optimization on hierarchical domains.
We demonstrate its effectiveness on complex system design problems, including a neural network and a green-aircraft case study.
Our methods are available in the open-source Surrogate Modeling Toolbox (SMT 2.0).}
}

\maketitle


\section*{Nomenclature}

\begin{table}[!h]
  \centering
  \label{tab:nomenclature}
  \small
  \begin{tabular}{@{}ll@{}}
    \toprule
    \textbf{Abbreviation} & \textbf{Meaning} \\
    \midrule
    ADSG      & Architecture Design Space Graph \\
    BO        & Bayesian Optimization \\
    cat & Categorical variable \\
    cont & Continuous variable \\
    CR & Continuous Relaxation \\
    DAG & Directed Acyclic Graph \\
    dec & Decreed variable \\
    DoE       & Design of Experiments \\
    DOT &  DAG of tomorrow \\
    DRAGON    &  Distributed fans Research Aircraft with electric Generators by ONERA \\
    DSG & Design Space Graph \\
    EI        & Expected Improvement \\
    EXC & Excluded \\
    EMI & Earth Mover Intersection \\
    FAST-OAD  & Future Aircraft Sizing Tool with Overall Aircraft Design  \\
    FM & Feature Model \\
    GED & Graph Edit Distance \\
    GD & Gower Distance \\
    HPO & Hyper-Parameters Optimization \\
    GP        & Gaussian Process \\
    HH & Homoscedastic Hypersphere \\
    int & Integer variable \\
    KPLS      & Kriging with Partial Least Squares \\
    LCA & Lowest Common Ancestor \\
    LHS       & Latin Hypercube Sampling \\
    met & Meta variable \\
    MAP & Maximum \textit{A Posteriori} \\
    MLP & Multi-Layer Perceptron \\
    MMD & Maximum Mean Discrepancy \\
    NSGA2     & Non‑dominated Sorting Genetic Algorithm II \\
    ord & Ordinal variable \\
    PLS       & Partial Least Squares \\
    qtn & Quantitative variable \\
    SBArchOpt & Surrogate-based Architecture Optimizer \\
    SEGOMOE   & Super Efficient Global Optimization with Mixture of Experts \\
    SMT       & Surrogate Modeling Toolbox \\
    SPD & Symmetric Positive Definite \\
    SVM & Singular Vector Machine \\
    UCB  & Upper Confidence Bound \\
    WB2S      & Watson and Barnes $2^{\mathrm{nd}}$‑criterion Scaled \\
    \bottomrule
  \end{tabular}
\end{table}


\section{Introduction} 
\label{sec:intro}

Optimizing the architecture of complex systems, such as those related to aerospace design, requires the exploration of vast and highly intricate design spaces to identify optimal solutions~\cite{bussemaker2022adore,chan2023goal}. These design processes often depend on \textit{expensive-to-evaluate} computer simulations, which are used to explore innovative concepts in early-stage studies and analyses~\cite{MartinsNing22}. 
\textcolor{black}{Given input values, these computer models output key performance metrics, such as the aerodynamic efficiency of an aircraft. Many of such models could be opaque, expensive-to-evaluate, or derivative-free and therefore are treated as \emph{blackbox} functions~\cite{AuHa2017}.}
The design and optimization of such complex systems is further complicated by the large number of variables influencing their overall performance and feasibility. As system complexity grows, traditional methods for exploring design spaces become less practical due to the high computational cost and inherent complexity of the data structures involved~\cite{MartinsNing22, Simmons2008}.
To address the inefficiencies of directly evaluating such simulations, the optimization process increasingly leverages \textit{surrogate models}, which provide inexpensive approaches to assess the expensive objective or constraint functions. These surrogate models enable the efficient exploration of complex and high-dimensional design spaces, making them indispensable for the optimization of complex systems~\cite{MartinsNing22}.

One of the critical challenges in this domain is the structured \textit{heterogeneity} of the design spaces, characterized by diverse input data types and complex interdependencies.
\textcolor{black}{
Real-world optimization problems may involve a combination of continuous, integer, and categorical variables. Moreover, some variables, referred to as \emph{decreed}, may be conditionally included or excluded depending on the state of other \emph{meta} variables, thereby introducing hierarchical structure into the design space~\cite{halle2024graph}.}
Consequently, such mixed-variable domains pose significant challenges that traditional optimization and modeling techniques often struggle to capture in a scalable and efficient manner, especially when the design space is structured and the relationships between variables are non-trivial~\cite{bhosekar2018advances,herrera}.
Although the literature offers several approaches for tackling these challenges, ranging from Bayesian optimization and surrogate modeling to feature-based methods, no single solution has been consistently outperforming~\cite{Talbi}.

As a result, the field lacks a comprehensive framework that can unify these varied techniques and address the full spectrum of mixed-variable and hierarchical optimization problems encountered in complex system design. This paper also addresses a critical gap between the SMT library~\cite{SMT2019} and the Design Space Graph (DSG) software (adsg-core)~\cite{bussemaker2024ADSG,bussemaker2025system}, which is essential for fully integrating hierarchical surrogate modeling into the existing ecosystem.
Our goal is to define distances and models for heterogeneous data points in mixed-variable domains, where two points do not necessarily share the same variables or bounds~\cite{halle2024graph}.
Such a domain is equipped with a single graph whose purpose is to model the complex hierarchical dependencies.
Although closely related, the proposed framework does not compare graphs, such as in graph matching.
The framework focuses on the variables that can be present across the different data configurations.
However, note that graph distances are covered in the literature review, since some notions of these are integrated in the framework\textcolor{black}{,} and they share similar purposes. 
\textcolor{black}{In consequence, our framework provides a consistent mathematical foundation for representing and comparing hierarchical, conditional, and heterogeneous variable domains, enabling their direct use within surrogate models such as Gaussian processes, inverse distance weighting, or $k$-nearest neighbors. It can also structure the input space of original simulations, serving as a formal interface between architecture definition and model evaluation.}

This work is motivated by various real-world applications where meta variables play a pivotal role. For example, thermal insulation design involves meta variables that dictate the number of heat intercepts, each introducing new design variables~\cite{Abra04, KoAuDe01a}.    
Similarly, in aerospace design, meta variables influence critical decisions such as the number of engine shafts or the inclusion of a fan~\cite{BuCiDeNaLa2021, bartoli2023Agile, bussemaker2024system}. 
\textcolor{black}{We position this work as a unified representation and modeling framework for mixed-variable, variable-size design spaces, primarily intended to support surrogate-based optimization of expensive, derivative-free simulators. While model-agnostic, the framework is not intended to replace methods that operate on completed graph objects (for example, graph-matching or specialized graph kernels) when structural isomorphism is the central modeling requirement; in such cases, complementary graph-centric techniques are preferable~\cite{perez2024gaussian}.}

The general context of this research lies in system architecture, which defines how systems meet stakeholder requirements by specifying components, functions, and their interrelationships~\cite{crawley2015system}. Designing system architectures involves complex decision-making, requiring cross-disciplinary collaboration and effective management of requirements and functions~\cite{chan2023goal, roussel2023assembly}. Our approach is validated through a case study on green aircraft design optimization using Bayesian Optimization (BO) techniques, demonstrating the framework's ability to manage hierarchical dependencies and complex variable structures. 
This paper builds upon prior work connecting hierarchical variable frameworks with surrogate-based modeling and optimization~\cite{halle2024graph, saves2024system,bussemaker2024system}, aiming to integrate these frameworks within more general system architecture environments~\cite{bussemaker2022adore, bussemaker2024ADSG, chan2022trying}. 
\textcolor{black}{Importantly, the proposed distance of~\cite{halle2024graph} is not merely a technical device but a practical enabler for surrogate modeling and optimization in mixed-variable, hierarchical domains.
It provides a principled way to compare two heterogeneous design points that may not share the same active variables.
Because of this, whole datasets can be used without partitioning by active-variable pattern.
The distance also permits construction of valid correlation kernels for Gaussian-process surrogates over variable-size domains.
This enables information transfer across subspaces, improving sample efficiency and making surrogate-based optimization (e.g., Bayesian optimization) more reliable.
The distance was developed and validated in our prior work~\cite{halle2024graph}, where additional experiments and implementation details are provided.}

This paper provides a comprehensive review of multiple research domains, identifying critical intersections and synergies in the existing literature. Then, this paper proposes a unified framework for surrogate modeling in architecture optimization, building on these findings. 
Leveraging \textit{graph theory} concepts, its goal is to extend existing concepts through a generalized approach capable of handling mixed-variable and hierarchical blackbox functions and simulations. To finish, this paper delivers the first open-source implementation of the developed framework, enabling the resolution of diverse heterogeneous modeling problems in practical applications. Our main contributions are as follows:
\begin{enumerate}
\setlength\itemsep{0cm}
    \item Conduct an extensive review of graph kernels and hierarchical variable models to contextualize their applications and to showcase their applicability in managing hierarchical dependencies and complex variable structures, particularly for defining distances or correlation kernels.

    \item 
    Propose a unified mathematical framework enabling more expressive and flexible modeling of dependencies \textcolor{black}{to integrate them within surrogate modeling workflows.} 
    This generalization provides a robust foundation for addressing hierarchical and conditional variable relationships. A distance is proposed to compare two hierarchical input points and build a correlation kernel for surrogate modeling. 

    \item  Implement this framework in the open-source SMT toolbox~\cite{saves2023smt}, extending its capabilities to support hierarchical surrogate models. This implementation is documented in a dedicated tutorial notebook\footnote{\url{https://github.com/SMTorg/smt-design-space-ext/blob/main/tutorial/SMT_DesignSpace_example.ipynb}}, easing its practical application.
        \item  \textcolor{black}{Showing the application of our method on two different test cases: (i) a neural network hyperparameter modeling problem to illustrate the design space modeling capabilities and (ii) a green aircraft multidisciplinary problem to illustrate the surrogate model capabilities for optimization purposes.}
\end{enumerate}

The remainder of this document is structured as follows.  
Section~\ref{sec:intro_literature_review} reviews related work, focusing on surrogate modeling in variable-size spaces, representations of structured data, and general distance measures over such structures.  
Section~\ref{sec:framework_ext} introduces our extended framework, detailing the hierarchical distance and kernel, the automation of the underlying graph construction, and demonstrating its use through a case study on multi-layer perceptron architectures.  
\textcolor{black}{Section~\ref{sec:application} applies the framework to a neural network model and extends it to surrogate emulation for the Bayesian optimization of a green aircraft.}  
Finally, Sect.~\ref{sec:conclu} summarizes our contributions and outlines directions for future research.

\section{Literature review}
\label{sec:intro_literature_review}
This section provides a literature review of related works and previous studies. Specifically, it reviews key approaches in handling variable-size spaces, data representations, and distance measures over complex structures.

\subsection{Background and previous works}

This section describes meta and meta-decreed variables, which model hierarchical dependencies in domains of interest, as introduced in previous works~\cite{audet2022general,MScEdward} and~\cite{halle2024graph}, respectively.
This framework has been formally analyzed through a so-called design space graph introduced for an application to hybrid-electric distributed propulsion aircraft~\cite{Mixed_Paul_PLS} and compared with the classical Feature Model (FM)~\cite{benavides2010automated}.

\subsubsection{Types and roles of variables}
\label{sec:role_graph}
For a given problem featuring $n$ variables, a point $\bold{X}$ can be decomposed as $\bold{X} = ({X}_1, \ldots, {X}_n)$, where ${X}_i$ is the value taken by the point $\bold{X}$ in its corresponding $i^\text{th}$ variable.
To incorporate structure into the problem, it is necessary to establish a terminology that describes the types and roles of variables. Each variable is assigned a \textit{type} that reflects its quantitative or qualitative nature~\cite{Mixed_Paul,bussemaker2024system}, four types that can be assigned to a variable:  
\begin{itemize}
\setlength\itemsep{0cm}
\item continuous, if the variable can take any real value between lower and upper bounds;
\item integer, if the variable can take any integer value between lower and upper bounds;
\item ordinal, if the variable can take any value in an ordered finite set of either strings or values; 
\item categorical, if the variable can take any value in an unordered finite set of either strings, modalities, or values. 
\end{itemize}
\textcolor{black}{
Concerning variables that take values in a finite set—namely \emph{integer}, \emph{ordinal}, and \emph{categorical} variables—the key distinction is the presence (or absence) of an intrinsic ordering among their levels. Categorical (nominal) variables have \emph{no} natural order: for example, a variable \textit{shape} with levels \{\text{square}, \text{circle}, \text{star}\} is unordered. By contrast, integer and ordinal variables are ordered. An integer variable carries a known, equally spaced order (\textit{e.g.}, \{1, 2, 3\}), whereas an ordinal variable admits an order but the spacing between successive levels need not be equal or even known (for instance \{1, 2, 4\} has non-uniform spacing, while \{\text{small}, \text{medium}, \text{large}\} is ordered but the distances between levels are qualitative). When the metric structure of an ordinal scale is unknown, one may learn a monotone embedding or distance function from data; for example, the authors of~\cite{Roustant} propose methods to estimate a strictly monotone mapping for ordinal levels. 
}

In addition to their variable type, each variable is assigned a \textit{role} that reflects its
hierarchy with other variables in points of the hierarchical domain~\cite{audet2022general, halle2024graph}.
An underlying graph structure models the hierarchy between variables through parent-child relations.
In this graph, a variable $x_i$ is a parent of another variable $x_j$ (a child), if $x_i$ controls the inclusion or bounds of $x_j$, with $i,j \in \{1,2,\ldots,n\}$.
In such cases, $x_j$ is said to have a \textit{decree dependency} with $x_i$, 
and an arc from $x_i$ to $x_j$ is present in the graph.
This arc represents the decree dependency.
More details on the graph structure are provided in the next section.
The roles of the variables, \textit{variable roles} for short, are determined based on whether they have parents and/or children in the graph.
Each variable is assigned one of the following four roles:  
\begin{itemize}
\setlength\itemsep{0cm}
\item (strictly) meta, if it has no parent and at least one child;
\item meta-decreed, if it has at least one parent and at least one child;
\item (strictly) decreed, if it has at least one parent and no children; 
\item neutral, it has neither parents nor children.
\end{itemize}
\textcolor{black}{
This taxonomy with four roles (meta, meta-decreed, decreed, neutral) captures the structural position of each input within the role graph and has direct modeling consequences. Intuitively, a \emph{meta} variable is a top-level controller (no parent, at least one child) whose values determine which downstream variables are admissible; a \emph{decreed} variable is a child-only input (at least one parent, no children) whose admissible set depends entirely on its parents; a \emph{meta-decreed} variable plays both roles (it is controlled by parents and in turn controls children); and a \emph{neutral} variable is independent of the hierarchical structure (no parents, no children). Concretely, these roles determine whether a variable can be included or excluded. Role assignment also guides dimensionality reduction and problem decomposition (for example, excluded decreed variables are omitted from surrogate evaluations and acquisition computations), and it supports interpretable modeling choices that respect the semantics provided by domain experts~\cite{halle2024graph, audet2022general}. Notwithstanding, this work is limited as it does not consider partially decreed variables, that is, the variables that are included but can only take a subset of their admissible values. Also, this previous work cannot handle incompatibilities properly, and the next sections will show how we can extend the previous work to generalize it.}
\subsubsection{Modeling Design Spaces featuring Graph Structures}
\label{sec:het_data}
This section presents references addressing the general class of problems that involve hierarchical spaces in the context of data-driven regression and optimization.
In this study, a hierarchical space, originally introduced in~\cite{HuOs2013}, is defined as a space with at least one meta variable, following the formalization in~\cite{halle2024graph}.
A hierarchical space may also be referred to as a variable-size space~\cite{pelamattihier}, as a conditional search space problem~\cite{Lucas_b}, as \textcolor{black}{a} heterogeneous dataset~\cite{remadi2024prompt}, or as a hierarchical domain~\cite{halle2024graph}.
More details about other data structure representations are given in Appendix~\ref{sec:data_representation}.

Decreed and meta-decreed variables may have multiple parents, and both types can influence numerous other variables. Building on these observations, we introduce the role graph \textcolor{black}{that is} a structure that captures all information about variable roles and decree dependencies.
Formally, the role graph $G=(N,A)$ is a couple, where $N$ is the set of all the variables represented as nodes of the problems and $A$ is the set of decree dependencies represented as arcs.

In~\cite{halle2024graph}, a general class of mixed-variable hierarchical structures within domains was introduced. This work rigorously defines a hierarchical domain, denoted as $\mathcal{X}_G$, that represents a domain $\mathcal{X}$ equipped with a graph structure $G$. In this domain, each point $\bold{X}$ is properly defined within $\mathcal{X}$, and the variables are hierarchically ordered in $G$, where a variable is a child of another if it is decreed by it. Equivalently, the domain is \textit{hierarchical} if its corresponding role graph contains at least one decreed variable (\textit{i.e.}, a variable with the decree property).
Additionally, the hierarchical domain framework introduced in~\cite{halle2024graph} models mixed-variable domains with meta variables, implying that points do not necessarily share the variables or bounds. 
This hierarchical domain $\mathcal{X}_G$ can be extended to contain the variables that can be considered excluded, \textit{i.e.}, the variables decreed excluded due to the values of the corresponding meta variables at a given point.

This extension yields a space, called the \textit{extended domain}, containing all the possible variables that are present in at least one point.
The role graph contains the decree dependencies and allows for conditioning universal sets into restricted sets.
It is assumed to be directed and acyclic to avoid pathological situations where meta-decreed variables influence each other.
Nevertheless, the Design Space Graph (DSG)  addresses this issue by parsing the design variables through a recursive process, stepping through the included (but not-yet-parsed) choices. 
Typically, parsers process design variables in a sequential, left-to-right manner. However, due to the complex interdependencies among design choices \textcolor{black}{in systems engineering like Architecture Design Space Graph (ADSG)}, a simple sequential parsing approach may not suffice. To address this, the DSG employs a recursive parsing strategy. This method allows the parser to navigate through the network of design variables, processing each based on its dependencies rather than its position in a sequence. 
This recursive approach ensures that all design variables, including those not initially parsed due to their position or dependencies, are eventually analyzed, leading to a more comprehensive understanding of the system's design structure~\cite{bussemaker2025system}. 

\begin{figure}[h!]
\centering
    \scalebox{0.6}{
    
    \begin{tikzpicture}

\node (wings) at (0,0) {};

\node(motors)[label={\large meta}, xshift=0.325cm, yshift=0cm, 
at=(wings), draw, ellipse]
{\begin{tabular}{c}
\# of  motors \\ 
$m \in \{0,1,2\}$
\end{tabular}};

\node(propellers1)[label={\large meta-decreed}, xshift=6.1cm, yshift=2cm, at=(motors), draw, ellipse]
{\begin{tabular}{c}
\# of propellers \\ 
for the first extra motor 
$p_{1} \in \{1,2\}$
\end{tabular}};
\draw[->] (motors)--(propellers1) node[midway, sloped, anchor=center, above] {};
\node(propellers2)[label={}, xshift=6.1cm, yshift=-2cm, at=(motors), draw, ellipse]
{\begin{tabular}{c}
\# of propellers \\ 
for the second extra motor 
$p_{2} \in \{1,2\}$
\end{tabular}};
\draw[->] (motors)--(propellers2) node[midway, sloped, anchor=center, above] {};


\node(radius11)[label={\large decreed}, xshift=10.1cm, yshift=1cm, at=(propellers1), draw, ellipse]
{\begin{tabular}{c}
Radius of the first propeller \\ 
in the first motor 
$r_{1,1} \in \mathbb{R}$
\end{tabular}};
\draw[->] (propellers1)--(radius11.west) node[midway, sloped, anchor=center, above] {};
\node(radius12)[label={}, xshift=10.1cm, yshift=-1cm, at=(propellers1), draw, ellipse]
{\begin{tabular}{c}
Radius of the second propeller \\ 
in the first motor
$r_{1,2} \in \mathbb{R}$
\end{tabular}};
\draw[->] (propellers1)--(radius12.west) node[midway, sloped, anchor=center, above] {};

\node(radius21)[label={}, xshift=10.1cm, yshift=1cm, at=(propellers2), draw, ellipse]
{\begin{tabular}{c}
Radius of the first propeller \\ 
in the second motor 
$r_{2,1} \in \mathbb{R}$
\end{tabular}};
\draw[->] (propellers2)--(radius21.west) node[midway, sloped, anchor=center, above] {};
\node(radius22)[label={}, xshift=10.1cm, yshift=-1cm, at=(propellers2), draw, ellipse]
{\begin{tabular}{c}
Radius of the second propeller \\ 
in the second motor 
$r_{2,2} \in \mathbb{R}$
\end{tabular}};
\draw[->] (propellers2)--(radius22.west) node[midway, sloped, anchor=center, above] {};


\end{tikzpicture}
    
    }
\caption{Variables roles for aircraft design problem.}
 \label{fig:intro_example2}
\end{figure}
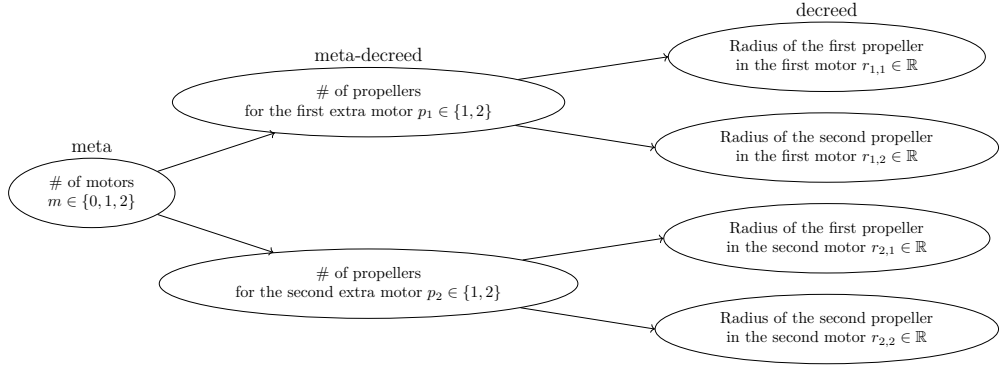

The \texttt{DRAGON} aircraft concept ~\cite{schmollgruber}
relies on a distributed electric propulsion aircraft, improving fuel efficiency by optimizing propulsive performance. \texttt{DRAGON} introduces multiple compact electric fans on the wing pressure side, increasing the bypass ratio as an alternative to large turbofans. This design overcomes challenges of under-wing turbofans while enabling transonic flight. 
In such aircraft design examples, a meta-decreed variable could be the number of motors that is used for a given wing, where each motor has a set of decreed \textcolor{black}{variables}.  
Figure~\ref{fig:intro_example2}~illustrates the roles of variables for a toy example of a wing aircraft design with meta-decreed variables.  
The number of motors per wing  $m$ is a meta variable, since it controls how many variables, peculiar to each motor, are in the problem ($m=0$ means single-engined aircraft, as the engine is not supported by the wing).
The number of propellers $p_i$ assigned to the $i$-th motor is a meta-decreed variable since 1)~its inclusion depends on the number of motors $m$, \textit{i.e.}, it is decreed, and 2)~it controls the number of variables regarding the radii of every included propeller. 
Finally, the radius $r_{ij}$ of the $j$-th propeller in the $i$-th motor is a decreed variable, since its inclusion depends on the number of propellers $p_{i}$ for the motor~$i$. 
In Fig.~\ref{fig:intro_example2}, to compute the meta distance, we consider the extended points that contain all included and excluded variables. The variables that are excluded for a given point are assigned the special value $\texttt{EXC}$.
In other words, some variables may be excluded for a point $\bold{X} \in \mathcal{X}$ but present in another point $\bold{X'}$.
For example, in Fig.~\ref{fig:intro_example2}, the variable representing the number of propellers of the second motor is set to $m_2=\textsc{exc}$ when the number $m$ of motors is one.
Considering the excluded variables provides the necessary information for computing distances between points that do not share the same variables. This trick is always possible by adding bounds for every variable: for example, if we only have data for an aircraft with 2 motors and 2 propellers, we can always restrict the extended design space without loss of generality. 
Consequently, the extended domain, in which extended points reside, is constructed directly with Cartesian products on the bounded restricted sets of each variable. 
The hierarchical distance is defined on the extended domain, but fortunately, a bijection based on the role graph $G$ enables the mapping of points to extended points by adding \texttt{EXC} to the excluded variables, and conversely, maps extended points back to original points by removing variables assigned the value \texttt{EXC}.
Thus, a distance function that considers all included and excluded variables is also induced on the original domain within the restricted set. For example, one such distance combines angular and length information, and is well-defined for heterogeneous data comparison~\cite{halle2024graph}. As such, it has the property of leading to unit induced distances when comparing an included variable's value with a non-included variable whose value is \texttt{EXC}. This unit distance is effective for multiplication-based comparisons, as it will be non-influential if multiplying the remaining distances by 1.   

\subsubsection{Hierarchical spaces with feature models}
\label{sec:fm_intro}

To describe complex configuration spaces—where selecting a high-level option determines the availability of related sub-options—we rely on \emph{feature models}, a formalism widely used in software engineering~\cite{bruel2020comparing}. In this context, a \emph{feature} represents a configurable aspect of a system, such as a module, parameter, or capability that can be enabled or disabled. This directly corresponds to the notion of variables in hierarchical modeling, where the activation of one variable governs the inclusion or structure of others~\cite{benavides2010automated}. For example, in aircraft design, the feature "{Energy source}" might offer alternatives such as "{Electricity}" or "{Fuel}", each unlocking its own specific set of design parameters. In this sense, features may correspond to variable levels, variables themselves, or even the domains (supports) of those variables. While feature models often define these roles loosely, we adopt a more structured interpretation in this work by treating features as the \textit{nodes} of a DSG, a formalism we build upon in subsequent sections~\cite{asadi2016goal, bussemaker2024ADSG}.
In a feature model, features are organized via a tree structure, where nodes represent features and edges define dependencies. A child feature is only accessible if its parent is selected. Some are \emph{mandatory}—always included when the parent is included—while others are \emph{optional}. Groups of child features can also form exclusive (XOR) or inclusive (OR) sets, expressing constraints like “select exactly one” or “select at least one.”
In addition to the tree, \emph{cross-tree constraints} allow logical rules between features in different branches. These include relations such as “A requires B” or “A excludes B”, as well as more complex logical relationships, as in “A and B implies not C”~\cite{batory2005feature}. 
Feature models provide a compact way to describe rich, structured design spaces without enumerating every valid configuration. In our setting, they naturally map to hierarchical variable spaces, where high-level meta-choices conditionally activate relevant subsets of parameters.

\subsection{Hierarchical surrogate modeling}

This section first reviews surrogate modeling strategies for representing architectural decision variables. It then tackles the core challenge of defining valid distance metrics in these complex, non-standard design spaces.

\begin{figure}[h!]
\centering
    \includegraphics[width=0.7\textwidth]{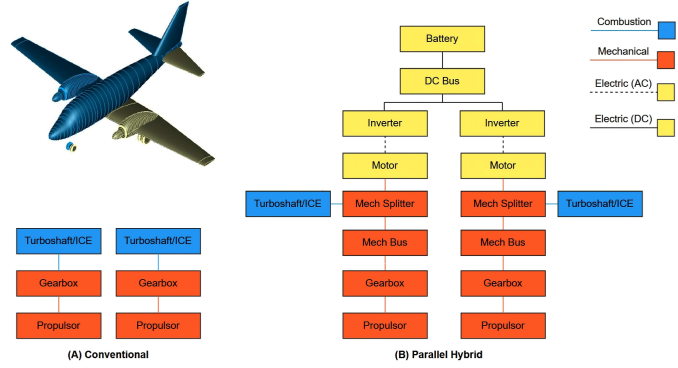}
\caption{\textcolor{black}{An example of a hierarchical surrogate in the context of aircraft design. Adapted from~\cite[Figure 4]{bussemaker2023function}.}}
 \label{fig:aircraft}
\end{figure}

\textcolor{black}{
The motivation for developing a hierarchical surrogate arises from the need to accurately model and optimize complex engineering architectures such as hybrid-electric propulsion systems, where multiple physical domains (mechanical, electrical, and thermal) interact. Traditional flat representations fail to capture these nested dependencies and conditional relationships between subsystems. As illustrated in Fig.~\ref{fig:aircraft}, the authors applied a function-based architecture modeling approach to the design of a hybrid-electric propulsion system for the Beechcraft King Air C90GT aircraft. Their framework systematically generated and evaluated thousands of propulsion architectures under mission constraints (\textit{e.g.}, climb, cruise, descent) and subsystem-level design variables (\textit{e.g.}, motor rating, battery mass, gearbox ratio). Such systems naturally combine discrete architectural choices like power-split topology and component inclusion with continuous design variables, leading to heterogeneous and nested design spaces that challenge traditional surrogate modeling. The present work therefore aims to construct surrogate models and distance metrics that preserve this hierarchical structure, enabling efficient learning, comparison, and optimization across large and multi-level configuration spaces~\cite{bussemaker2023function}.
}

\subsubsection{Hierarchical design spaces for surrogate modeling}

Interest in modeling "variable-size" spaces emerged from various disciplines around 2010. For example, the work by~\cite{HuJoMe2009} introduced Latin Hypercube Sampling (LHS) and Gaussian process (or Kriging) for \textit{nested} variables, where variables depend hierarchically on other variables. Although the mechanisms differ, these nested variables can be considered akin to decreed variables in this context.
Similarly, the technical report~\cite{HuOs2013} from 2013 proposes a mixed-variable kernel function for varying-size spaces, referred \textcolor{black}{to} as hierarchical.
The latter report has been a key reference for the previous work~\cite{halle2024graph}\textcolor{black}{;} thus, it is also important for this current work. In this report~\cite{HuOs2013}, the inclusion and exclusion of a variable is controlled by variable-wise Kronecker delta functions.
\textcolor{black}{This variable-wise Kronecker delta denotes a per-variable indicator of mutual inclusion: only variables active in both sample points contribute to their kernel term. This idea resonates with the kernel designs in variable-size Bayesian optimization~\cite{Pelamatti2020}, which allow covariance across samples with partially distinct dimensional supports ~\cite{pelamattihier}.}
More precisely, the domain is equipped with a Directed Acyclic Graph (DAG) structure that contains hierarchical dependencies between the variables and where the variables are viewed as nodes.
The inclusion of a (child) variable is determined by a delta function that takes the values of its ancestors, that is, variables of higher hierarchy that share a path in the DAG.  The root and intermediate nodes are restricted to categorical variables. The kernel is constructed variable-wise with multiplication and addition of one-dimensional kernel.
Each one-dimensional kernel is computed with an assigned isometry, which is a distance-preserving function that maps a variable into a Euclidean space~\cite{feragen2015geodesic}.
The computation of a one-dimensional kernel is divided into three cases: the variable is included for both points or the variable is excluded for both points, or the variable is included for one and only one point.

In a Bayesian optimization context, the paper~\cite{PeBrBaTaGu2021} proposes kernel functions that allow the construction of a GP on domains with dimensional variables~\cite{LuPi04a, LuPiSc05a}, which are a special case of strictly discrete meta variables that influence the dimension, and propose two kernels. First, the subproblem-wise kernel computes similarity measures between dimensional variables, as well as between the other included variables, whose inclusion can be determined by the dimensional variables.
Second, the dimensional variable-wise kernel decomposes the computation dimensional variable per dimensional variable, which allows \textcolor{black}{for regrouping} the specific decreed variables that are included by their corresponding dimensional variables~\cite{HuJoMe2009}. 
For SVM regression, the work~\cite{GaDuKaSe2018} proposes the Earth Mover’s Intersection (EMI or discrete Wasserstein) that computes similarity measures between sets of different sizes, similarly to the Jaccard index~\cite{tanimoto1958elementary}.
A kernel function is induced from the EMI, and SVM regression experiments are realized.

Going back to GP regression for hierarchical design spaces, defining valid and efficient covariance kernels over unordered sets of vectors presents unique challenges.
Traditional kernels, like the squared exponential kernel, often fall short when applied to sets with no inherent ordering and varying cardinalities.
Methods such as the Maximum Mean Discrepancy (MMD) and the Wasserstein distance offer promising solutions by treating these sets as samples from distributions, enabling the comparison of sets with different sizes and structures~\cite{fellmann2024kernel}.
For instance, the MMD-based kernel has shown strong performance in adapting to the geometric properties of underlying functions, making it a robust choice for surrogate modeling in complex engineering problems~\cite{sow2023learning}.
However, such methods are not structure-specific and may struggle with high-dimensional data or scenarios with significant missing information, underscoring the need for more specialized kernels that can effectively capture the underlying structure and relationships within the data.

In the complex systems architecting context, the works~\cite{ bussemaker2024ADSG,bussemaker2025system,BuCiNa2020} detail the Architecture Design Space Graph (ADSG) and its application for modeling complex architectural \textcolor{black}{design} spaces with a semantic approach. In this approach, three types of architectural decisions are available: function-component mapping, component characterization, and component connection. The graph model is constructed from a design
space definition, and discrete architectural decisions are automatically inserted according to
specified rules. The DSG abstracts away from the system architecture context, and uses a directed graph to model general-purpose hierarchical selection choices (\textit{i.e.}, selections between mutually-exclusive options) and connection choices (\textit{i.e.}, connection patterns between sets of source and target nodes). 
The current work integrates the architectural decisions of the DSG, notably for the automatic construction of the domain in the software part of this work.
However, the DSG supports cyclic graphs and therefore includes, but is not limited to, DAGs. 
A notable tool in this domain is \texttt{ConfigSpace}~\cite{SMAC3}, an open-source Python library that facilitates the definition and management of hierarchical design spaces. \texttt{ConfigSpace} supports mixed-discrete variables, conditional activation of design variables, and value constraints. It allows for querying and validating design vectors, correcting and imputing design vectors, and generating valid design vectors. It is widely used by \textcolor{black}{surrogate} modeling and optimization frameworks such as SMAC3~\cite{SMAC3}, BOAH~\cite{Lindauer2019}, OpenBox~\cite{jiang2023openbox}, and SMT~\cite{saves2023smt}. Even if explicit hierarchical models are not available, problems often contain some implicit hierarchical structure~\cite{bussemaker2021system}. Various design space modeling techniques, including the Architecture Decision Graph~\cite{Simmons2008}, the Adaptive Reconfigurable Matrix of Alternatives~\cite{Mavris2008}, or the \textcolor{black}{aforementioned} DSG, to cite a few, provide useful frameworks for understanding and optimizing complex systems. 
More relevant background on data representations within structured spaces is given in Appedix~\ref{sec:data_representation}.

\subsubsection{General distances over structures} 
\label{sec:distances}
One of the objectives of this paper is to propose a distance to compare two hierarchical input points and to build a correlation kernel on top of such a distance. The goal of this section is to give more insights into these particular domains. 

To begin with, recall 
that a function $d$ is called a distance if and only if $d$ respects the four criteria of positivity (1), identity (2), symmetry (3), and the triangular inequality (4). 

In particular, when a function only respects the points 1, 3, and 4, it is called a pseudo-distance, and when it only respects the points 1, 2, and 4, it is called a quasi-distance~\cite{deza2014other}. 

Another closely related concept is the similarity $s$~\cite{goshtasby2012similarity} defined as a function that respects the boundness (instead of positivity), identity, symmetry, and respects the triangular inequality.
However, several authors do not include the symmetry in the definition of the similarity measure~\cite{tversky1979studies}, while some authors add the positivity but remove the triangular inequality~\cite{ontanon2020overview}. Furthermore, some authors introduce dissimilarity measures that respect only positivity and reflexivity (instead of identity) and divergence measures that respect both positivity and identity~\cite{da2015global}. 
Therefore, distances must be preferred over similarity distances. Most definitions of similarity measures include correlation, and, for example, Pearson's R correlation~\cite{bravais1844analyse} is bounded, symmetric, and respects the identity, but not the positivity and the triangular inequality~\cite{chen2023triangle}. This is the case with most correlation similarities (\textit{e.g.}, distance correlation, Rho-vectorial coefficient~\cite{robert1976unifying}, Brownian covariance~\cite{szekely2009brownian}, polychloric correlations~\cite{coenders1995categorization}).

Both distances and similarity measures can be distinguished according to their deterministic or stochastic nature. For example, distances between vectors include Minkowski and Mahalanobis distances, generalizing Euclidean distances~\cite{busemann1947isoperimetric} in, respectively, a deterministic and a stochastic setup. 
General objects could also be compared with respect to the precise features they have. For example, the cosine similarity (Otsuka-Ochiai coefficient)~\cite{steck2024cosine} is a measure between vectors with respect to the angle that they form. 
Distances tend to form families of distance and to rely on parameters and thus are connected to the fields of hyperparameter optimization, feature selection, and metric learning. More details about common distances are given hereinafter. 

In the previous section, we mentioned the EMI~\cite{GaDuKaSe2018} measure that extends the Jaccard similarity index, and the Tanimoto measure~\cite{friedli2025crps,gomez2018automatic} to compare generic variable-size vectors or deterministic sets~\cite{valls2016error}. 
Similar literature worth noting includes several basic distances over non-structured set. Among \textcolor{black}{others,} we can cite the Hamming distance~\cite{Gower}, Morisita's overlap index~\cite{butturi2020edge}, Szymkiewicz-Simpson coefficient~\cite{sager1976classification}, Renkonen similarity index~\cite{renkonen1938statisch}, Dice-Sørensen index~\cite{sorensen1948method}, Lee-Mannheim distance~\cite{nishimura2008generalization}, or 
Tversky index~\cite{tversky1979studies}.

Distances between strings or sequences can be measured using various probabilistic or deterministic methods. For time-series, deterministic approaches include resampling followed by Minkowski or cosine distances~\cite{keogh2002need}, frequency-based features like Fourier transforms~\cite{agrawal1993efficient}, and elastic measures such as time-warped edit distance~\cite{marteau2014recursive}. Stochastic approaches often rely on auto-regressive models~\cite{serra2014empirical}, such as using Kullback-Leibler divergence between Markov chains as a Bayesian similarity measure~\cite{ramoni2002bayesian}, or the Wasserstein-Fourier distance for stationary series~\cite{cazelles2020wasserstein}. For ordered data, rank-based metrics like Kendall’s Tau and Spearman’s Rho~\cite{kendall1948rank} compare element ranks instead of values. More generally, information-based measures like Kolmogorov complexity can be approximated by the normalized compression distance~\cite{cilibrasi2005clustering}. Simple distances can also be extended to hierarchical data, from sets~\cite{balancca2012set} to more complex structures discussed in the next section.

\subsubsection{Graph‐based distances to compare structured data}
\label{sec:ged}
Graph representation is among the most popular hierarchical database \textcolor{black}{representations,} even if not
the only solution to model such hierarchies. However, it is the \textcolor{black}{most} appropriate to our context of hierarchical variables for architectural problems (see Appendix~\ref{sec:data_representation} for more background information). 
In fact, rather than comparing individual decision variables, these methods treat each architecture as a complete graph and quantify similarity through operations such as edge insertion, deletion, and permutation‐based alignment of shortest‐path or adjacency matrices. By focusing on holistic graph topology, this alternative approach captures structural differences and commonalities in hierarchical designs, complementing variable‐level metrics with a well-established graph similarity perspective.

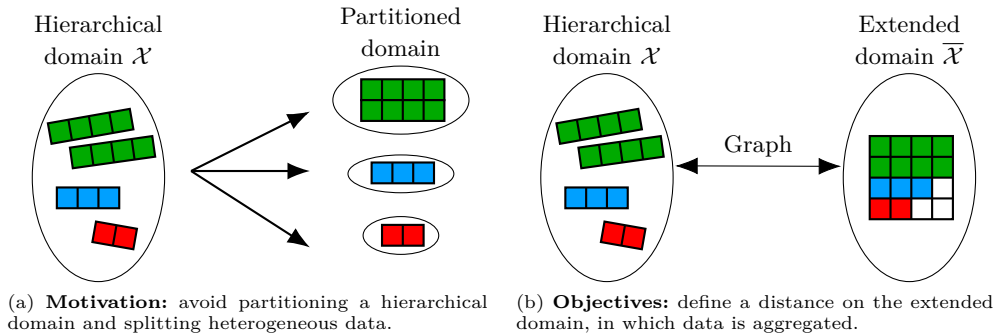
\begin{figure}[htb!]
\centering

\subfloat[\textcolor{black}{\textbf{Motivation:} avoid partitioning a hierarchical domain and splitting heterogeneous data.}\label{subfig:intro_big_picture_motivation}]{
  \scalebox{1}{
\begin{tikzpicture}

 \color{black}
    \definecolor{myyellow}{RGB}{0,170,0}
    \definecolor{mygreen}{RGB}{0,170,0}
    \definecolor{myblue}{RGB}{0,160,255}
    \definecolor{myred}{RGB}{255,0,0}

    \node[label={[align=center]\small {Hierarchical} \\ \small {domain} $\mathcal{X}$}, ellipse, draw, minimum width=1.75cm, minimum height=2.75cm] (Domain) at (5,0) {};

    \begin{scope}[shift={(4.45,-0.35)}, scale=0.275] 
        \begin{scope}[rotate=10, shift={(0.25,2.9)}]
            \fill[myyellow] (0,0) rectangle (4,1);
            \draw[thick] (0,0) rectangle (4,1);
            \foreach \x in {1,2,3} \draw[thick] (\x,0) -- (\x,1);
        \end{scope}

        \begin{scope}[rotate=8, shift={(1,1.6)}]
            \fill[mygreen] (0,0) rectangle (4,1);
            \draw[thick] (0,0) rectangle (4,1);
            \foreach \x in {1,2,3} \draw[thick] (\x,0) -- (\x,1);
        \end{scope}

        \begin{scope}[shift={(0,-0.2)}]
            \fill[myblue] (0,0) rectangle (3,1);
            \draw[thick] (0,0) rectangle (3,1);
            \foreach \x in {1,2} \draw[thick] (\x,0) -- (\x,1);
        \end{scope}

        \begin{scope}[rotate=-10, shift={(2,-1.5)}]
            \fill[myred] (0,0) rectangle (2,1);
            \draw[thick] (0,0) rectangle (2,1);
            \draw[thick] (1,0) -- (1,1);
        \end{scope}
    \end{scope}
    

    %
    \node[label={[align=center]\small {Partitioned} \\ \small {domain}}, ellipse,  minimum width=1.75cm, minimum height=2.75cm] (Extended) at (9,0.1) {};

    \begin{scope}[shift={(-1, 0)}]


        \begin{scope}[shift={(9.475, -0.55+0.2)}, scale=0.275]
            \fill[myyellow] (0,5) rectangle (4,6);
            \draw[thick] (0,5) rectangle (4,6);
            \foreach \x in {1,2,3} \draw[thick] (\x,5) -- (\x,6);
    
            \fill[mygreen] (0,4) rectangle (4,5);
            \draw[thick] (0,4) rectangle (4,5);
            \foreach \x in {1,2,3} \draw[thick] (\x,4) -- (\x,5);
    
            \begin{scope}[shift={(0.5,0)}]
                \fill[myblue] (0,1) rectangle (3,2);
                \draw[thick] (0,1) rectangle (3,2);
                \foreach \x in {1,2} \draw[thick] (\x,1) -- (\x,2);
            \end{scope}
            
            \begin{scope}[shift={(1,-2)}]
                \fill[myred] (0,0) rectangle (2,1);
                \draw[thick] (0,0) rectangle (2,1);
                \draw[thick] (1,0) -- (1,1);
            \end{scope}

        \end{scope}
          
    \end{scope}

    \begin{scope}[shift={(4.2, -0.25)}, scale=0.45]
         \draw [thick, -{Latex[length=3mm]}] (4.5, 0.75) -- (8, 2.5);  
        \draw [thick, -{Latex[length=3mm]}] (4.5, 0.75) -- (8, 0.75);  
         \draw [thick, -{Latex[length=3mm]}] (4.5, 0.75) -- (8, -1.5);  
    \end{scope}

    \begin{scope}[shift={(0.5, 0)}]
        \draw (8.5, 0.825+0.2) ellipse [x radius=0.9cm, y radius=0.45cm];
    
        \draw (8.5, -0.15+0.2) ellipse [x radius=0.7cm, y radius=0.25cm];

        \draw (8.5, -0.965+0.2) ellipse [x radius=0.5cm, y radius=0.25cm];
    \end{scope}
 

\end{tikzpicture}  
  }
}
\hspace{0.25cm}
\subfloat[\textcolor{black}{\textbf{Objectives:} define a distance on the extended domain, in which data is aggregated.}\label{subfig:intro_big_picture_approach}]{
  \scalebox{1}{
\begin{tikzpicture}
 \color{black}
    \definecolor{myyellow}{RGB}{0,170,0}
    \definecolor{mygreen}{RGB}{0,170,0}
    \definecolor{myblue}{RGB}{0,160,255}
    \definecolor{myred}{RGB}{255,0,0}
    \definecolor{mygray}{RGB}{255,255,255}
    \definecolor{mywhite}{RGB}{255,255,255}


    \node[label={[align=center]\small {Hierarchical} \\ \small {domain} $\mathcal{X}$}, ellipse, draw, minimum width=1.75cm, minimum height=2.75cm] (Domain) at (5,0) {};

    \begin{scope}[shift={(4.45,-0.35)}, scale=0.275] 
        \begin{scope}[rotate=10, shift={(0.25,2.9)}]
            \fill[myyellow] (0,0) rectangle (4,1);
            \draw[thick] (0,0) rectangle (4,1);
            \foreach \x in {1,2,3} \draw[thick] (\x,0) -- (\x,1);
        \end{scope}

        \begin{scope}[rotate=8, shift={(1,1.6)}]
            \fill[mygreen] (0,0) rectangle (4,1);
            \draw[thick] (0,0) rectangle (4,1);
            \foreach \x in {1,2,3} \draw[thick] (\x,0) -- (\x,1);
        \end{scope}

        \begin{scope}[shift={(0,-0.2)}]
            \fill[myblue] (0,0) rectangle (3,1);
            \draw[thick] (0,0) rectangle (3,1);
            \foreach \x in {1,2} \draw[thick] (\x,0) -- (\x,1);
        \end{scope}

        \begin{scope}[rotate=-10, shift={(2,-1.5)}]
            \fill[myred] (0,0) rectangle (2,1);
            \draw[thick] (0,0) rectangle (2,1);
            \draw[thick] (1,0) -- (1,1);
        \end{scope}
    \end{scope}


    \begin{scope}[shift={(-1, 0)}]


        \node[label={[align=center]\small {Extended} \\ \small {domain} $\overline{\mathcal{X}}$}, ellipse, draw, minimum width=1.75cm, minimum height=2.75cm] (Extended) at (10,0) {};

        \begin{scope}[shift={(9.475, -0.55)}, scale=0.275]
            \fill[myyellow] (0,3) rectangle (4,4);
            \fill[mygreen] (0,2) rectangle (4,3);
            \fill[myblue] (0,1) rectangle (3,2);
            \fill[mywhite] (3,1) rectangle (4,2);
            \fill[myred] (0,0) rectangle (2,1);
            \fill[mygray] (2,0) rectangle (4,1);
    
            \draw[thick] (0,0) rectangle (4,4);
            
            \draw[thick] (0,3) -- (4,3);
            \draw[thick] (0,2) -- (4,2);
            \draw[thick] (0,1) -- (4,1);
    
            \draw[thick] (1,0) -- (1,4);
            \draw[thick] (2,0) -- (2,4);
            \draw[thick] (3,0) -- (3,4);
        \end{scope}
          
    \end{scope}

    %


    \draw [{Latex[length=3mm]}-{Latex[length=3mm]}] 
    (5.9, 0.155) to [out=0, in=180] 
    node[midway, above] {\small Graph} 
    (7.6+0.5, 0.15);


\end{tikzpicture}  
  }
}

\caption{\textcolor{black}{Motivation for developing a unified distance. Adapted from~\cite[Figure~2]{halle2024graph}.}}
\label{fig:partition}
\end{figure}

\textcolor{black}{A central difficulty when working with hierarchical, variable-size domains is that two design points need not share the same active variables or the same dimensional support.
Figure~\ref{fig:partition} is intended as a conceptual motivation rather than a numerical illustration. It highlights why a distance is required when building surrogate models over hierarchical, variable-size design spaces: two designs may not share the same active variables or dimensional support. By embedding all designs into a common extended hierarchical domain, a structured distance can be defined to quantify similarity despite heterogeneous architectures. This distance is then used to construct a surrogate model in a hierarchical space, allowing a single surrogate to be trained on mixed and heterogeneous data. Continuous design variables contribute through standard metric terms, while discrete and conditional variables are handled through the graph-structured support, enabling statistically efficient modeling and optimization across architectures.
A common approach is to partition the hierarchical domain into homogeneous subspaces and treat each subspace separately. Figure~\ref{subfig:intro_big_picture_motivation} depicts a hierarchical design domain $\mathcal{X}$ in which different design points activate different subsets of mixed variables (shown by colored blocks). This leads to heterogeneous dimensional supports that include various numbers of continuous or discrete variables: for example, the first two variables are always included in every subproblem. Notably, this approach sacrifices statistical efficiency (data are split) and complicates downstream surrogate modeling and optimization because it fragments the data and hinders efficient space characterization for the models. }

 \textcolor{black}{The latter motivates our approach that embeds all hierarchical points into a single \emph{extended} domain $\bar{\mathcal{X}}$ (right), where inactive or excluded variables are explicitly represented so all data can be used jointly as in Fig.~\ref{subfig:intro_big_picture_approach}.  
Concretely, we construct an extended representation that embeds each hierarchical point into a common extended hierarchical space while retaining the notion of a variable's admissible \emph{support} (active, partially active, or excluded). The distance is then computed as a structured aggregation of per-variable  (as opposed to per-subspace) contributions that account for the graph-structured space.
This unified distance directly defines the correlation kernel used by the surrogate model, enabling learning and optimization across heterogeneous hierarchical designs without partitioning.
The latter distance allows for the construction of a surrogate model that admits practical scalability strategies when exhaustive pairwise computations become expensive.
}

Various methods measure similarity or distance between elements in hierarchies and taxonomies, categorized by feature-based~\cite{ontanon2012similarity}, morphism-based~\cite{wallis2001graph}, or structural properties, including deterministic and stochastic distances.

Deterministic distances, like Rada’s distance~\cite{rada1989development}, compute the distance in a hierarchy by counting edges up to the Lowest Common Ancestor (LCA). The LCA distance~\cite{bender2005lowest} measures the path length between two nodes starting from their common ancestor, with the idea that nodes with a more specific common ancestor are more similar. Path distance also fits into this category, as it measures the length of the path between two nodes, often considering their depth or level in the hierarchy~\cite{winter1992path}. Rada’s distance is defined as:
\begin{equation}
d_\text{Rada}(\bold{X},\bold{X'}) = \text{path}(\bold{X}, \text{LCA}(\bold{X}, \bold{X'})) + \text{path}(\bold{X'}, \text{LCA}(\bold{X}, \bold{X'})),
\end{equation}
where $ \text{LCA}(\bold{X}, \bold{Y}) $ represents the lowest common ancestor of $\bold{x}$ and $\bold{y}$ in the hierarchy.

Stochastic distances, such as Resnik’s distance~\cite{resnik1999semantic}, introduce the use of information content (IC) to measure similarity, based on the probability of encountering concepts and their hierarchical significance. Resnik’s distance is defined as:
\begin{equation}
d_\text{Resnik}(\bold{X}, \bold{X'}) = -\log \left( p(\text{LCA}(\bold{X}, \bold{X'})) \right).
\end{equation}
 Hybrid distances, \textit{e.g.}, Jiang-Conrath distance~\cite{jiang1997semantic}, combine edge counting and IC for a more comprehensive measure. 
Additionally, in the realm of stochastic approaches, Wasserstein-based distances, like the Sliced Wasserstein Weisfeiler-Lehman (SW-WL) kernel~\cite{perez2024gaussian}, provide a powerful framework for measuring structural similarity between meshes. These distances are commonly used in transport theory to compare distributions over graph features. These approaches are also useful when working with non-attributed graphs and enable the efficient computation of graph distances using multi-level refinements of graph structure~\cite{nikolentzos2021graph}.

Graphs are powerful tools for representing hierarchical structures, especially for heterogeneous data. Graph Edit Distance (GED)~\cite{sanfeliu1983distance} is widely used to quantify the similarity between two graphs by calculating the minimal cost to transform one graph into another through edit operations like node and edge insertions, deletions, or substitutions. GED generalizes several similarity measures, such as Hamming and Jaro-Winkler distances~\cite{winkler2014matching}. It satisfies properties like non-negativity, identity, symmetry, and the triangle inequality~\cite{serratosa2021redefining}.
The computation of GED depends on the type of graph considered. For attributed graphs, similarity is based on node and edge attributes~\cite{blumenthal2020exact}, with approaches including self-organizing maps~\cite{neuhaus2005self}, and graph kernels~\cite{neuhaus2006convolution}. For non-attributed graphs like directed acyclic graphs or trees, GED can be computed using string-based methods, such as tree edit distance, which takes advantage of polynomial-time tree isomorphism problems~\cite{fernandez2001graph}. Methods like Dijkstra’s algorithm~\cite{robles2004string} are often more efficient for these cases. These methods convert graphs into strings and compute the shortest edit path using algorithms such as Dijkstra’s or maximum \textit{a posteriori} (MAP)~\cite{robles2005graph}, which avoid matrix normalization to reduce complexity.

In addition to graph edit distance, one can measure similarity by aligning global matrix representations of the graphs.  Let \(A_\bold{X}\) and \(A_\bold{X'}\) be the adjacency matrices of graphs \(G_\bold{X}\) and \(G_\bold{X'}\), respectively, and let \(D_\bold{X}\) and \(D_\bold{X'}\) denote their corresponding shortest‐path distance matrices.  A family of permutation‐based distances seeks a permutation matrix \(P\) that best aligns these representations under the Frobenius norm.  For instance, the \emph{chemical distance}~\cite{kvasnivcka1991reaction} is
\begin{equation}
d^{P_n}_{\mathrm{Chem}}(G_\bold{X}, G_\bold{X'})
\;=\;
\min_{P \in P_n}\|\,A_\bold{X}\,P - P\,A_\bold{X'}\|_F,
\end{equation}

where \(P_n\) is the set of all \(n\times n\) permutation matrices \textcolor{black}{over the $n$ vertices of the considered graphs} and \(\|\cdot\|_F\) denotes the Frobenius norm, selecting the node alignment that minimizes adjacency discrepancies.
Likewise, the Chartrand-Kubiki-Shultz (CKS) distance~\cite{chartrand1998graph} applies the same principle to shortest‐path information,
\begin{equation}
d^{P_n}_{\mathrm{CKS}}(G_\bold{X}, G_\bold{X'})
\;=\;
\min_{P \in P_n}\|\,D_\bold{X}\,P - P\,D_\bold{X'}\|_F,
\end{equation}
emphasizing global connectivity differences.  Other related measures include Poole’s most interesting common generalization distance~\cite{poole1995novel}, which abstracts both graphs to their least general common ancestor before measuring edit cost, and the Champin-Solnon mapping cost~\cite{champin2003measuring}, which allows flexible (non‐bijective) vertex correspondences via a cost matrix.  Wang and Ishii’s formulation~\cite{wang1997method} unifies both adjacency‐ and path‐based distances under the same permutation framework, highlighting the versatility of Frobenius norm alignment for capturing holistic graph similarity. Still, these distances are related to the graph isomorphism problem~\cite{grohe2020graph}, making their computation challenging.

This review introduced key concepts and relevant works, identifying a critical gap: the lack of a unified framework or software for handling mixed-variable and hierarchical challenges in complex system design, such as architecture design.
The following section shows how to extend and generalize further the frameworks introduced in this section with application to surrogate modeling. 
    \section{A unified framework for hierarchical design spaces}
\label{sec:framework_ext}
\textcolor{black}{
Building on the literature reviewed in Sect.~\ref{sec:intro_literature_review}, our framework unifies feature models, mixed-variable kernels, and graph-based similarity.
It uses a design space graph representation that encodes conditional inclusion and partially-decreed relationships.
The representation supports continuous, integer, and categorical data and yields distances and kernels tailored for surrogate modeling.
In that sense, the present work generalizes prior mixed-variable and hierarchical kernels, \textit{e.g.}, imputation and role-agnostic strategies, while remaining complementary to full graph-matching and graph-kernel methods when explicit topology matching is required, \textit{e.g.}, Gaussian process over meshes~\cite{perez2024gaussian}.}

This section presents a unified graph-based framework to model structured design spaces with mixed variables, hierarchical dependencies, and custom distances, enabling the handling of structured data representations in architectural modeling and optimization~\cite{bussemaker2024ADSG,bihanic2012models}. Specifically, excluded variables offer valuable information for computing similarity measures between points that do not share the same variables. This approach enables hierarchical and mixed discrete variables to be effectively incorporated into a correlation kernel or distance metric, facilitating comparisons between architectural configurations.
Such correlation kernels belong to the family of graph-induced kernels~\cite{ramachandram2018bayesian}, which are frequently employed in Bayesian Optimization (BO)~\cite{le2021revisiting,sheikh2022bayesian} for heterogeneous datasets, making them a powerful tool for managing complex, structured design spaces~\cite{well-adapted_cont}.

\subsection{Adding relationships in-between variables}
In this section, we introduce the new relationships between the variables and their levels in the graph-structured design space, that is a hierarchical domain equipped with a graph structure modeling the hierarchical dependencies between the variables. In particular, we introduce these three new relationships.

\begin{figure}[h!]
\centering
    \scalebox{0.6}{
\begin{tikzpicture}

\node (wings) at (0,0) {};

\node(motors)[label={\large meta}, xshift=0.325cm, yshift=0cm, 
at=(wings), draw, ellipse]
{\begin{tabular}{c}
\# of  motors \\ 
$m \in \{0,1,2\}$
\end{tabular}};

\node(propellers1)[label={\large fully decreed}, xshift=10.1cm, yshift=4cm, at=(motors), draw, ellipse]
{\begin{tabular}{c}
\# of propellers \\ 
for the first extra motor 
$p_{1} \in \{1,2\}$
\end{tabular}};
\draw[->] (motors)--(propellers1) node[midway, sloped, anchor=center, above] {decree};

\node(propellers2)[label= {\large fully decreed}, xshift=10.1cm, yshift=0cm, at=(motors), draw, ellipse]
{\begin{tabular}{c}
\# of propellers \\ 
for the second extra motor 
$p_{2} \in \{1,2\}$
\end{tabular}};
\draw[->] (motors)--(propellers2) node[midway, sloped, anchor=center, above] {decree};

\node(length)[label={\large partially-decreed}, xshift=10.1cm, yshift=-4cm, at=(motors), draw, ellipse]
{\begin{tabular}{c}
Length of wings \\ 
$l \in [m+1, 4]$
\end{tabular}};
\draw[->] (motors)--(length) node[midway, sloped, anchor=center, above] {decree};


\end{tikzpicture}

    }
\caption{Variables roles for an aircraft design problem with a varying number of motors.}
 \label{fig:intro_example7}
\end{figure}
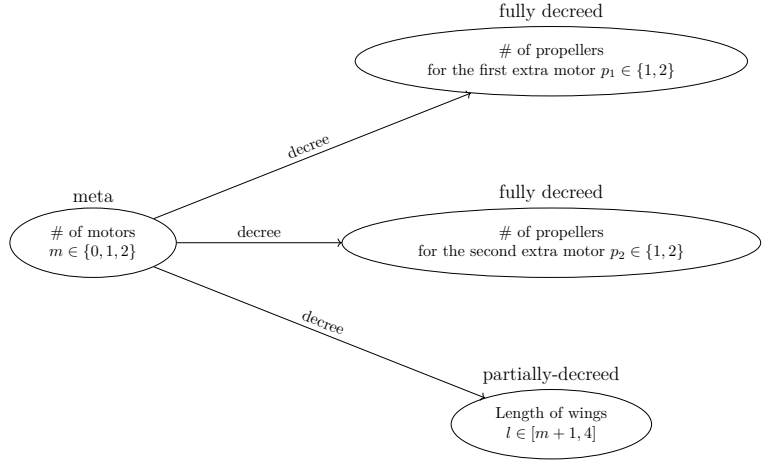

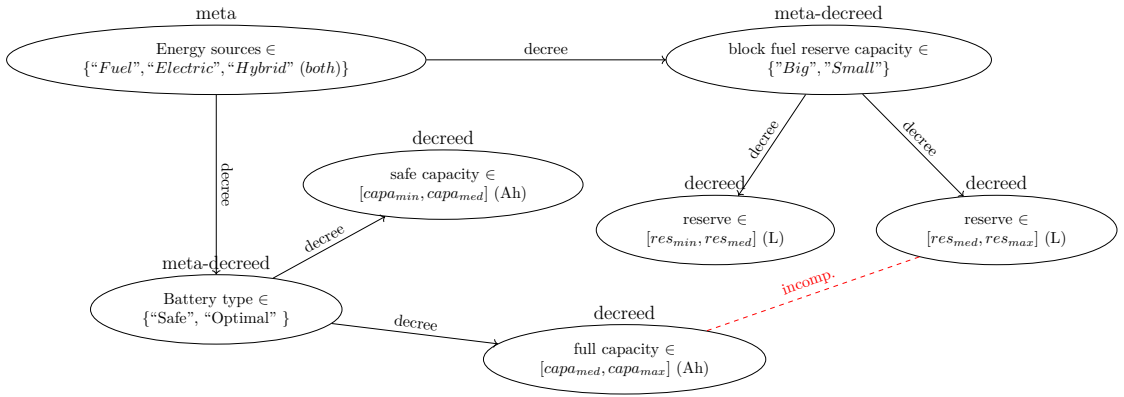
\begin{figure}[H]
\centering
    \scalebox{0.6}{

\begin{tikzpicture}

\node(sources)[label={\large meta}, draw, ellipse] at (0,0)
{\begin{tabular}{c}
$\text{Energy sources} \in $\\
$ \{ ``Fuel",``Electric",``Hybrid" \ (both) \} $  
\end{tabular}} ;

\node(assignments)[label={\large meta-decreed }, xshift=13.5cm, yshift=0cm, draw, ellipse]
{\begin{tabular}{c}
 block fuel  reserve capacity $\in$  \\
$\{"Big", "Small" \} $   
\end{tabular}};
\draw[->] (sources)--(assignments) node[midway, sloped, anchor=center, above] { decree};


\node(battery)[label={\large meta-decreed}, xshift=0cm, yshift=-5.5cm, draw, ellipse]
{\begin{tabular}{c}
 Battery type $\in$  \\
 \{"Safe", "Optimal" \}

\end{tabular}} ;
\draw[->] (sources)--(battery) node[midway, sloped, anchor=center, above] { decree};

\node(inter1)[label={\large decreed}, xshift=11cm, yshift=-3.75cm, draw, ellipse]
{\begin{tabular}{c}
reserve  $\in$  \\
$[res_{min},res_{med}] $  (L)
\end{tabular}} ;

\node(inter2)[label={\large decreed}, xshift=17.2cm, yshift=-3.75cm, draw, ellipse]
{\begin{tabular}{c}
reserve  $\in$  \\
$[res_{med},res_{max}] $  (L)
\end{tabular}} ;

\node(inter3)[label={\large decreed}, xshift=5cm, yshift=-2.75cm, draw, ellipse]
{\begin{tabular}{c}
safe capacity  $\in$  \\
$[capa_{min},capa_{med}] $  (Ah)
\end{tabular}} ;

\node(inter4)[label={\large decreed}, xshift=9cm, yshift=-6.6cm, draw, ellipse]
{\begin{tabular}{c}
full capacity  $\in$  \\
$[capa_{med},capa_{max}] $  (Ah)
\end{tabular}} ;

\draw[->] (assignments)--(inter1) node[midway, sloped, anchor=center, above] { decree};

\draw[->] (assignments)--(inter2) node[midway, sloped, anchor=center, above] { decree};

\draw[->] (battery)--(inter3) node[midway, sloped, anchor=center, above] { decree};

\draw[->] (battery)--(inter4) node[midway, sloped, anchor=center, above] { decree};

\draw[draw=red, fill=red, dashed] 
  (inter2) -- (inter4) 
  node[midway, sloped, above] {\textcolor{red}{incomp.}};



\end{tikzpicture}

    }
\caption{Variables roles for an aircraft design problem with a varying energy source.}
 \label{fig:intro_example8}
\end{figure}

\paragraph{Decree dependencies}  
This is the most general meta to decreed relationship between any two variables, capturing multi‐level and nested inclusion or exclusion dependencies as detailed in Sect~\ref{sec:role_graph}. To be more precise, we introduce \emph{partially‑decreed} variables for when a parent does not fully enable or disable its child, but rather restricts the child’s admissible values without entirely excluding it.
For the toy example in Fig.~\ref{fig:intro_example7}, the number of motors per wing  $m$ is a meta variable parent of the number of propellers $p_1$ and $p_2$, since it controls their inclusion or exclusion.
The variable $m$ is also a meta variable parent of the length of wings $l$ since it determines its admissible values.
The variable $l$ is partially-decreed, since it is always included, but its admissible values are controlled by $m$.
%
\paragraph{Incompatibility} An incompatibility relationship, similar to the “Excludes” constraint in feature models (Sect.~\ref{sec:fm_intro}), prevents two variables from taking certain combinations of values. For instance, a specific battery type may be incompatible with high fuel levels for hybrid energy due to safety concerns, as in Fig.~\ref{fig:intro_example8}. \textit{Mandatory} constraints can also be captured indirectly: if selecting a value excludes all but one option of another variable, the remaining option becomes implicitly required, effectively modeling a “Requires” relationship.
%

\begin{figure}[H]
\centering
    \scalebox{0.6}{
\begin{tikzpicture}

\node(sources)[label={\large meta}, draw, circle] at (0,0)
{\begin{tabular}{c}
$P_{max}  \in  [1.5,5] $ (bar) 
\end{tabular}} ;

\node(assignments)[label={\large meta-partially-decreed }, xshift=8.5cm, yshift=0cm, draw, circle]
{\begin{tabular}{c}
$P_{input}  \in  [1.5,P_{max}) $ (bar) 

\end{tabular}};
\draw[->] (sources)--(assignments) node[midway, sloped, anchor=center, above] {partially decree};


\node(end)[label={\large partially-decreed }, xshift=17cm, yshift=0cm, draw, circle]
{\begin{tabular}{c}
$P_{min}  \in  [1.5,P_{input}) $ (bar) 
\end{tabular}};
\draw[->] (assignments)--(end) node[midway, sloped, anchor=center, above] {partially decree};

\end{tikzpicture}    
    }
\caption{Roles of variables for a pressure order example.}
\label{fig:intro_example4}
\end{figure}
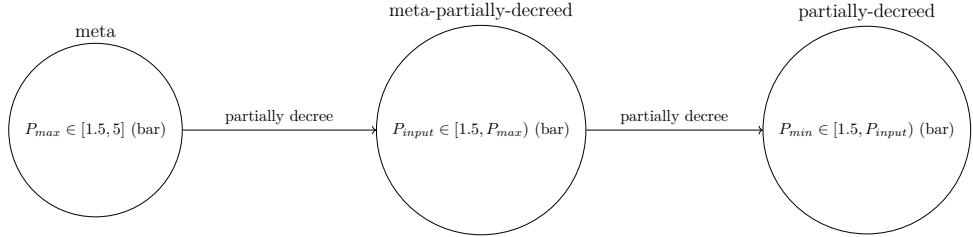

\paragraph{Order relationship} For continuous variables, whose supports cannot be enumerated, incompatibilities are expressed through order relations (\textit{e.g.}, inequalities). Equality constraints can, similarly to before, be constructed from the conjunction of two opposite inequalities, enabling \textit{Mandatory} relationships in a continuous setting.
For instance, in designing a hydrogen tank, if there are three pressure variables such that $P_{min} < P_{input} < P_{max}$~\cite{parello2024structural}, this relationship holds true with the structure of Fig.~\ref{fig:intro_example4} structure. 

\begin{figure}[H]
\centering
    \scalebox{0.6}{

\begin{tikzpicture}

\node(sources)[label={\large meta}, draw, ellipse] at (0,0)
{\begin{tabular}{c}
$\text{Energy sources} \in $\\
$ \{ ``Fuel",``Electric",``Hybrid" \} $  
\end{tabular}} ;

\node(assignments)[label={\large decreed }, xshift=15cm, yshift=0cm, draw, ellipse]
{\begin{tabular}{c}
 block fuel  reserve capacity $\in$  \\
$[reserve_{min},reserve_{max}] $  (L)
\end{tabular}};
\draw[->] (sources)--(assignments) node[midway, sloped, anchor=center, above] { decree};


\node(battery)[label={\large decreed}, xshift=0cm, yshift=-7.6cm, draw, ellipse]
{\begin{tabular}{c}
 Back-up battery capacity $\in$  \\
$[capa_{min},capa_{max}] $  (Ah)
\end{tabular}} ;
\draw[->] (sources)--(battery) node[midway, sloped, anchor=center, above] { decree};

\node(inter)[label={\large decreed}, xshift=8.2cm, yshift=-4.2cm, draw, rectangle]
{\begin{tabular}{c}
Intermediate var. $\in$  \\
$\{0,1\} $ s.t. \\
Inter. value $= \mathds{1}_{res_{capa} < res_{cut}} $
\end{tabular}} ;
\draw[->] (sources)--(inter) node[midway, sloped, anchor=center, above] { decree};
\draw[->] (assignments)--(inter) node[midway, sloped, anchor=center, above] {partially decree};
\draw[->] (inter)--(battery) node[midway, sloped, anchor=center, above] {partially decree};


\end{tikzpicture}

    }
\caption{Roles of variables for a hybrid-electric technology choice example.}
\label{fig:intro_example5}
\end{figure}
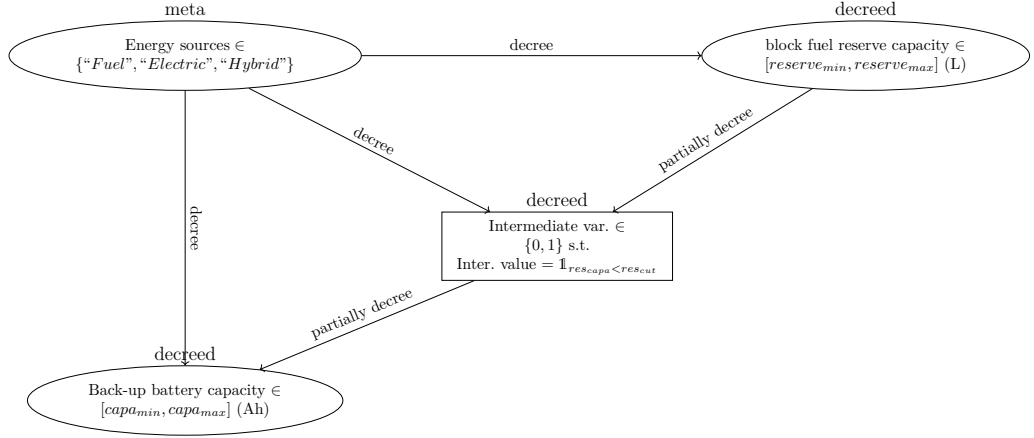

We can go even further by mixing categorical variables and partially-decreed continuous bounds.
A dummy example inspired by~\cite {fouda2022automated} is as follows.
The energy used for propulsion could be either "Fuel", "Electric" or "Hybrid" (both fuel and electricity). If the energy is either electric or hybrid, a back-up battery needs to be sized, but if the energy is hybrid and the fuel reserve is sufficient, no back-up battery is needed. Therefore, the back-up battery capacity is a decreed variable that is included if the energy is electric or if the energy is hybrid but the reserve is too small. To model the "and" interaction, an "intermediate node" as shown in Fig.~\ref{fig:intro_example5}, needs to be added to the model. The intermediate variable is decreed by the energy variable and is included only if the source is a hybrid electric, and its value is 1 if the fuel reserve is insufficient and 0 otherwise. The backup battery is included if the intermediate value is included and is of value 1 or if the energy source is electric. 

This work extends the notion of variable roles to accommodate a wider range of applications, including aerospace system design~\cite{BuCiDeNaLa2021, BuCiNa2020, Simmons2008} and software architecture~\cite{AlBuGrKoMe2013}. Unlike traditional feature models, our representation is a DAG rather than a tree, allowing features to have multiple parents. The framework is more formally described in the following section.
\subsection{Generalizing the hierarchical definition}
\label{sec:framework_extension}

Compared to the previous work~\cite{halle2024graph}, we extend the graph to include more concepts and have a structure that is more complex but more flexible and more general. This new structure adds the novel concepts developed in the previous section and is partly inspired by feature model and graph theory concepts. Consequently, it is not a plain directed graph $(N,A)$ but a so-called mixed graph $(N,A,E)$, that mixes arcs $A$ and undirected edges $E$. Notwithstanding, we limit our work to graphs that are directed and acyclic if we do not consider the undirected edges.

\begin{mydef}[Generalized role graph]
The generalized role graph $\mathcal{G} = \left(N, A, E \right)$ is a graph structure, where
\begin{itemize}[leftmargin=*,labelindent=16pt]

    \item $N$ is { the set of nodes} that contains 1) all the included and excluded variables, together with their levels if discrete or bounded supports if continuous, { and 2) all intermediate variables}, 
    
    \item $A$ is { the set of decree dependencies} that contains references for all inclusion-exclusion or admissible values dependencies between two nodes, represented as arcs,
    
   \item $E$ is the set of incompatibilities that contains references for all incompatibilities between two nodes, represented as edges. 
    
\end{itemize}

\noindent A node $n \in N$, which refers to any variable or support, can be of three possible kind: $N =  N_{\text{var}} \cup N_{\text{levels}} \cup  N_{\text{bnd}} \cup N_{\text{mid}}$, where $N_{\text{var}}$ is the set of possible variables, $N_\text{levels}$ is the set of the levels of any discrete variable, $ N_{\text{bnd}} $ is the set of continuous bounds for any continuous variable, and  $ N_{\text{mid}} $ is the set of intermediate nodes modeling composite (\textit{e.g.},\ “and/or”) activation conditions. Likewise, features in features model in Sect.~\ref{sec:fm_intro}, a node is more general and not limited to a variable.

\noindent An arc $a \in A$, which refers to any dependency, connects a parent (variable) to a child (variable), whose inclusion or admissible values are influenced by the parent or to a child (level/bounds) that is part of the complete support of the variable. We then have $A =  A_{\text{dec-met}} \cup A_{\text{bnd-lim}} 
$ where $ A_{\text{dec-met}}$ is the set of meta/decreed relationships, $A_{\text{bnd-lim}}$ is the set of partially meta/decreed relationships.

\noindent An edge $e \in E$, which refers to any incompatibility, can be of two kinds $ E = E_{\text{opposed}} \cup E_{\text{order}} $, where $E_{\text{opposed}}$ is the set of completely incompatible two nodes and  $E_{\text{order}} $ is the set of order relationships between two variables.
\label{def:role_graph}
\end{mydef}

As in Sect.~\ref{sec:role_graph}, a variable is assigned one type and one role. Their types are continuous (cont), ordinal (ord), integer (int),  or categorical (cat) as explained in Sect.~\ref{sec:role_graph}. In particular, ordinal variables are intermediate variables that can be either integer, such as \{2,4,8\} or categorical, such as \{small, medium, high\} but with an order relation (high$>$medium$>$small) and are therefore quantitative (qnt) like integer or continuous variables. Variables also come with bounds or support/levels definitions. The role is among meta $(\meta)$, meta-decreed $(\metadec)$, decreed $(\decreed)$, or neutral $(\neutral)$. 

In this framework, each node can have parents that are the other nodes in $N$ that directly influence its activation or admissible values. A node’s parents are simply those connected to it by a directed arc in the graph. These node parents must themselves be included (\textit{i.e.}, not excluded from the design).
To handle the heterogeneous design space, we extend each point to include all possible dimensions, enabling direct comparisons between them—a strategy formally validated in~\cite{halle2024graph}, which relies on the notion of support as a variable admissible values depend on its parent values. The support, knowing the role graph $\mathcal{G}$, is defined as follows:

\begin{mydef}[Support]
Let $x_i$ be a variable with full value space $\mathcal{X}_i$ and let $P_i \subset N$ be the set containing its parents in the graph $\mathcal{G}$. The support of $x_i$, that depends on the values taken by its parents, is written as $\mathcal{S}_{\mathcal{G}}(x_i \mid P_i) \subseteq \mathcal{X}_i$ or $\mathcal{S}_i \subseteq \mathcal{X}_i$ for short. This support is the set of admissible values that $x_i$ can take under the design space graph $\mathcal{G}$ and the values taken by its parents. A variable is considered \emph{excluded} from the design space when $\mathcal{S}_{\mathcal{G}}(x_i \mid P_i) = \emptyset$.
\label{def:partial_support_v2}
\end{mydef} 
\textcolor{black}{ Here, \textcolor{black}{the \emph{full value space} of a variable is the set of all possible values that this variable could take in the absence of any constraints from parent variables.} The support $\mathcal{S}_{\mathcal{G}}(x_i \mid P_i)$ captures the subset of these values that are admissible given the values of its parents in the design space graph $\mathcal{G}$. This distinction allows explicit modeling of excluded or partially-decreed variables.}
In fact, a variable could be excluded, and in that case its support is the empty set, but it also could be partially included, and in such case its support is a subset of the full support.
This leads to the following definition of a partially-decreed variable, a variable that is always present, but whose admissible values are constrained by the values of its parent.

\begin{mydef}[Partially-decreed variable]
A variable $x_i$ is said to be \emph{partially-decreed} by its parents $x_j \in P_i$, if
\[
\mathcal{S}_{\mathcal{G}}(x_i \mid x_j) \subset \mathcal{X}_i \quad \text{and} \quad \mathcal{S}_{\mathcal{G}}(x_i \mid x_j) \neq \emptyset.
\]
This generalizes the decree logic~\cite{audet2022general}: if $\mathcal{S}_{\mathcal{G}}(x_i \mid x_j) = \mathcal{X}_i$, then $x_i$ is fully included (conditionally included); if $\mathcal{S}_{\mathcal{G}}(x_i \mid x_j) = \emptyset$, it is excluded (conditionally excluded).
\label{def:partial_support_v3}
\end{mydef} 

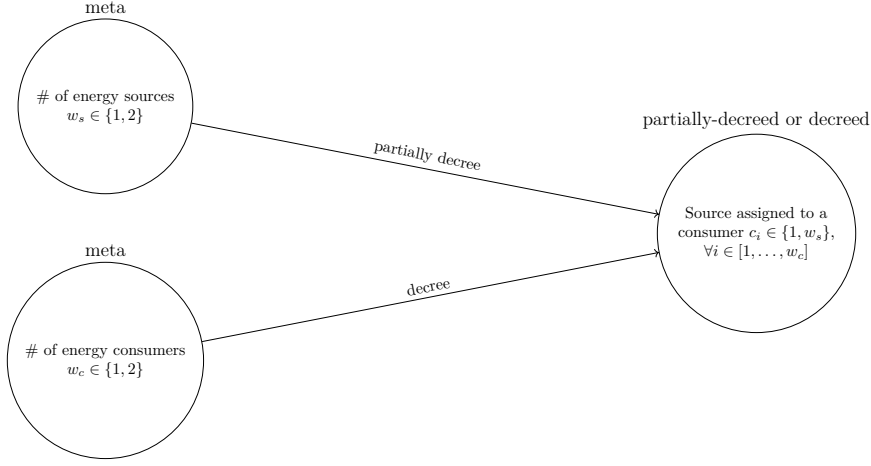
\begin{figure}[H]
\centering
    \scalebox{0.6}{
\begin{tikzpicture}

\node (wings) at (0,0) {};

\node(motors)[ xshift=4.25cm, yshift=0cm, 
at=(wings)] {};

\node(propellers)[xshift=5.1cm, yshift=0cm, at=(motors)] {};

\node(sources)[label={\large meta}, draw, circle] at (0,0)
{\begin{tabular}{c}
\# of energy sources \\ $w_{s} \in \{1,2\}$ 
\end{tabular}} ;

\node(consumers)[label={\large meta}, xshift=0cm, yshift=-5.6cm, draw, circle] at (0,0)
{\begin{tabular}{c}
\# of energy consumers \\ $w_{c} \in \{1,2\}$ 
\end{tabular}} ;

\node(assignments)[label={\large partially-decreed or decreed}, xshift=5.0cm, yshift=-2.8cm, at=(propellers), draw, circle]
{\begin{tabular}{c}
Source assigned to a\\ 
consumer $c_{i} \in  \{1,w_s\}$, \\
$\forall i \in [1,\ldots, w_c]$
\end{tabular}};
\draw[->] (consumers)--(assignments) node[midway, sloped, anchor=center, above] {decree};
\draw[->] (sources)--(assignments) node[midway, sloped, anchor=center, above] {partially decree};


\end{tikzpicture}    
    }
\caption{Roles of variables for a source-to-consumer example.}
\label{fig:intro_example3}
\end{figure}

The source-to-target assignment problem features such variables~\cite{bussemaker2024system}.
In that problem, either 1 or 2 consumers are assigned to either 1 or 2 energy sources. Two hierarchical interactions arise:
if there is only one source, then both consumers have to be assigned to that source, and if there is only one consumer, the source choice for the second consumer is not included.
Figure~\ref{fig:intro_example3}~illustrates the roles of variables for this source-to-target example.
This problem includes 4 variables: the number of energy sources between 1 or 2, the number of consumers between 1 or 2, the source of energy used by the first consumer, and the source of energy used by the second consumer.
There are 4 meta possibilities: if 1 source and 2 consumers, the assignment variables are partially-decreed because only the possibility "source number 1" can be chosen.
If 2 sources and 2 consumers, the assignment variables are fully decreed, both "source number 1" and "source number 2" are available for both consumers.
If 2 sources and 1 consumer,  the variable "source of energy for the first consumer" is fully decreed but the variable "source of energy for the second consumer" is completely excluded because there is no second consumer, in that case, the only possibility for that variable is "no possibilities" or empty set.
If 1 source and 1 consumer,  the variable "source of energy for the first consumer" is partially-decreed, but the variable "source of energy for the second consumer" is completely excluded.
As shown by this example, the same variable can be either partially or completely decreed depending on the problem and meta variable.
Therefore, contrarily to other roles, partially-decree and decree property are not mutually exclusive roles and as such, we may also consider meta-partially-decreed variables, that is variables that are both meta and partially-decreed. In particular, if a meta variable prohibits a value of another meta-partially-decreed variable, and if this value was decreeing another variable, then, the first meta variable is in fact excluding the latter variable: nested variable can create complex graph relations.

However, in our novel extended framework, it is possible to have completely incompatible variables and therefore we need to be able to compare two points without any choice-dependence. The latter remark excludes some classical approaches to compare heterogeneous data such as the imputation method\textcolor{black}{~\cite{bussemaker2024system,sportisse2020imputation}}.
Consequently, in the next section, we will show how to construct a well-behaved distance and kernel to deal with mixed hierarchical points. 

\subsection{Mixed hierarchical kernels for hierarchical domains}
\label{sec:GP-BO}

SMT 2.0 introduced a hierarchical correlation kernel that accounts for both meta and decreed variables, and was further extended in~\cite{saves24} to handle distances between decreed continuous variables. In this work, we generalize the SMT framework to support fully hierarchical domains. This extension enriches the \texttt{DesignSpace} definition with support for meta-decreed, partially-decreed, and order-based relationships, as described in this section\footnote{\url{https://smt.readthedocs.io/en/latest/_src_docs/applications/Mixed_Hier_usage.html}}. With these variable types and structural relations, we can now define a generalized design space suitable for hierarchical surrogate modeling.
\begin{mydef}[Graph-structured design space]
The design space $\mathcal{X}$ represents the set of all possible and { admissible} configurations of variables based on the graph $\mathcal{G}$.
Each point $\bold{X} \in \mathcal{X}$ is a configuration defined by a tuple of variable values $( {X}_1, {X}_2, \dots, {X}_n)$, with ${X}_i$ is the value of $\bold{X}$ for the $i$-th variable $x_i$,  and it may include continuous, discrete, or categorical variables.
\label{def:design_space}
\end{mydef}
\noindent The span of the design space is the Cartesian product of the supports of the variables possibly included within it, where we define a distance function
\begin{equation}
d_i : \mathcal{S}_i \times \mathcal{S}_i  \;\longrightarrow\; \mathbb{R}_{+},
\end{equation}
The function \(d_i( {X}_i, {X}_i')\) is defined as:
\begin{equation}
d_i({X}_i, {X}_i') =
\begin{cases}
    d(X_i, X_i'), & \text{if } X_i,X_i' \in \mathcal{S}_i , \\[6pt]
  \delta_i, & \text{if exactly one of } {
  X}_i, X_i' = \emptyset, \\[6pt]
  0, & \text{if } X_i = X_i' = \emptyset,
\end{cases}
\end{equation}
where $d$ is any 1-dimensional distance function (see Sect.~\ref{sec:distances}) and
where $\delta_i= \max\{ d(X_i, X_i') \, : \, X_i, X_i' \in \mathcal{S}_i \}/2 \geq 0$ is constant ensuring the triangular inequality is respected between included and excluded variables, as proven in~\cite{halle2024graph}.
 Note that a particular case of the proof for $\delta_i$ is given in Appendix~\ref{app:SPD}.
Finally, for any $p \geq 1$, the hierarchical function $\dist_p:\mathcal{X} \times \mathcal{X} \to \mathbb{R}^+$ defined by
\begin{equation}
     \dist_p( \bold{X}, \bold{X'}   ) := 
    \left( \sum\limits_{i=1}^{n}  \hspace{2pt} {d}_i \left( X_i, X'_i \right) ^p \right)^{\sfrac{1}{p}}.
\label{eq:graph_structured_distance} 
\end{equation}

In~\cite{gao2010survey}, the authors found good results with kernel approaches because "Unlike the traditional edit distance, kernel functions make good
use of statistical learning theory in the inner product rather than the graph space directly". Therefore, this work was motivated by the need to extend graph distances to handle mixed hierarchical variables.  \noindent We generalize the SMT 2.0 hierarchical kernel as the product of three Symmetric Positive Definite (SPD) sub‐kernels: a neutral kernel for variables without dependencies, a meta kernel for high‐level configuration choices, and a meta-decreed kernel that ties each choice to its activated children via our graph‐distance.  Because each factor is SPD, their product is SPD as well, and the resulting kernel naturally handles mixed variable types, conditional activations, and hierarchical structure. More details and the SPD proof are given in Appendix~\ref{app:SPD}.
Also, in this work, we will use the so-called algebraic distance introduced in~\cite{saves24}, namely 
\begin{equation}
d(\bold{X},\bold{X'}) = 
\begin{cases}
1 , & \parbox{3cm}{
$\text{if}\ \bold{X}^\top \bold{X'} = 0 $}  \\ 
\frac{ ||\bold{X}-\bold{X'}||}{ \sqrt{||\bold{X}||^2+1}\sqrt{||\bold{X'}||^2+1}}, & \text{otherwise.}
\end{cases}
\end{equation}
This distance is well-defined as proven in~\cite{halle2024graph} and the kernel that results from such distance, known as the Alg-Kernel, is SPD as proven in Appendix~\ref{app:SPD}. This distance is a particular case of the more general hierarchical distance introduced in~\cite{halle2024graph}.


\section{Modeling and optimizing with graph-structure inputs}
\label{sec:application}
In this section, we will connect the implementations of the mixed hierarchical surrogate models with the graph-structured design space. \textcolor{black}{In particular, we will describe the method implemented in SMT to build a hierarchical GP and show an illustration on two test cases. First, the {hyperparameters} modeling of a neural network and, second, an application to a green aircraft optimization through Bayesian optimization.
Section~\ref{sec:working_example_MLP} and Sect.~\ref{sec:ad} respectively introduce the two problems, that are MLP modeling and aircraft design modeling. Then, Sect.~\ref{sec:gsds} presents the graph-structured design space introduced in this work put into practice, and Sect.~\ref{sec:mlpapp} showcases its application to the MLP model. To finish with, Sect.~\ref{sec:bo} presents the practical details of the graph-structured Bayesian optimization, and Sect.~\ref{sec:boapp} showcases its application to the aircraft design model. 
}

\subsection{Modeling a complex MLP system architecture}
\label{sec:working_example_MLP}
\textcolor{black}{ To illustrate our approach, we introduce a complex system design problem with different variable types and dependencies.
For that, we leverage the Multi-Layer Perceptron (MLP) example that has been introduced in~\cite{audet2022general,halle2024graph,shihua2025bayesian}, where an MLP is tuned for a given supervised learning task by optimizing its hyperparameters. }
Identifying optimal hyperparameters is a problem called HyperParameter Optimization (HPO) that is known to be challenging~\cite{Goodfellow-et-al-2016}. In fact, the training, validation, and performance testing are typically conducted with a predetermined set of hyperparameters~\cite{hypernomad_paper}.
These hyperparameters vary widely in type, with some even acting as meta variables, such as the number of hidden layers which is a good illustration for a particular framework.
Gaussian Processes (GPs) offer valuable assistance in this optimization process.
GPs can efficiently evaluate multiple sets of hyperparameters at a significantly reduced computational cost.
Furthermore, GPs can be integrated into Bayesian optimization frameworks to automate the search for optimal hyperparameters~\cite{klein2017fast}.
The deep model's performance relative to its hyperparameters can be conceptualized as a blackbox function with mixed hierarchical variable inputs and therefore heterogeneous data.

\begin{figure}[!htb]
    \centering
  \includegraphics[width=\textwidth]{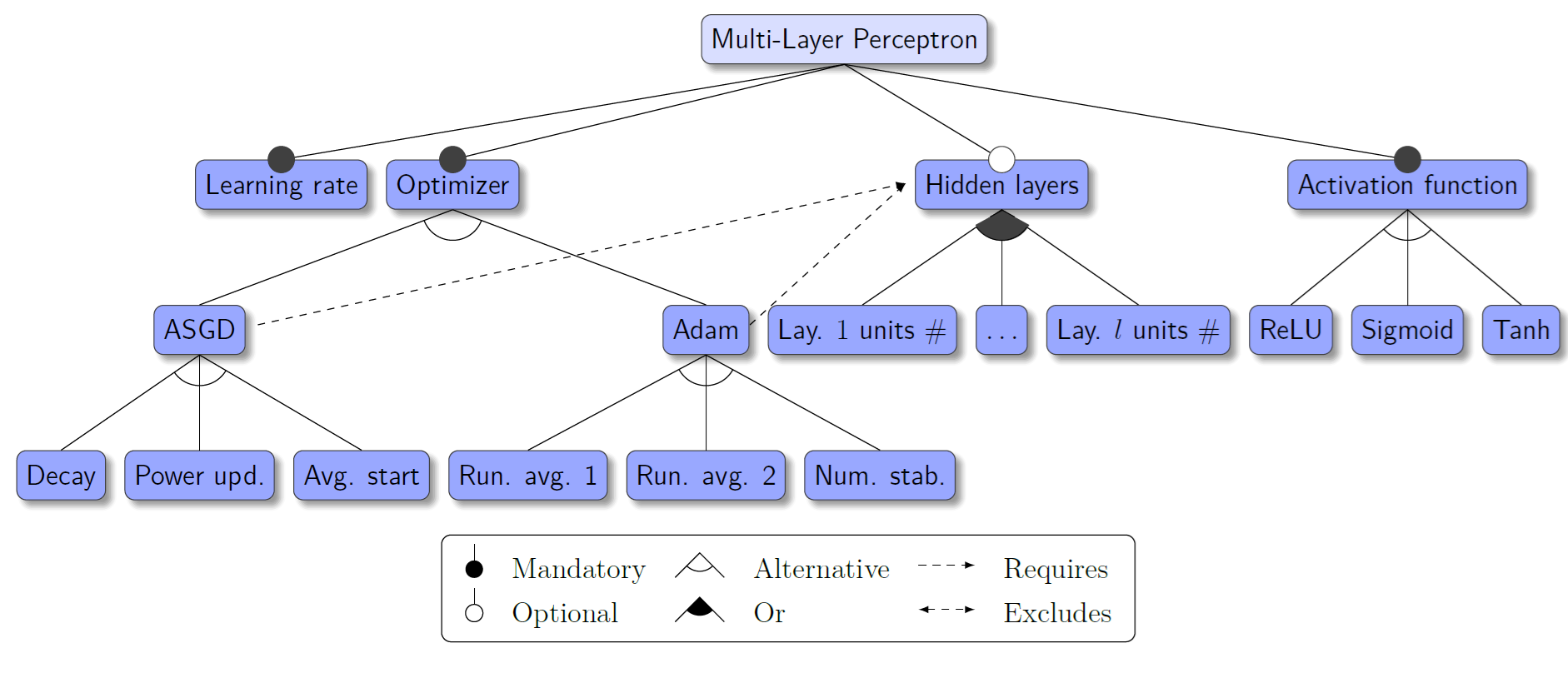}
    \caption{Feature model for the MLP working example.}
    \label{fig:MLP_FM}
\end{figure}

Consider a blackbox function $f:\mathcal{X} \to \mathbb{R}$, which outputs a performance score $f(\bold{X}) \in \mathbb{R}$ for a given vector of hyperparameters $\bold{X} \in \mathcal{X}$.
For the MLP problem, the function $f$ is evaluated by training the deep model knowing the specific set of hyperparameters $\bold{X}$ and then test it to obtain a resulting performance score $f(\bold{X})$. Therefore, the latter represents the model's accuracy on a previously unseen data set~\cite{audet2022general}.
While the internal mechanics of $f$ are understood, the function is treated as an expensive-to-evaluate blackbox problem due to the complexity involved in adjusting billions of model parameters and of fine-tuning them through backpropagation during training~\cite{audet2022general, hypernomad}.
To facilitate the comprehension of our modeling work, Feature Model (FM) has been used to build the tree presented in Fig.~\ref{fig:MLP_FM} that shows how modeling can give insights for a system such as the MLP. First, the selection of the optimizer determines which hyperparameters are included in the optimization domain.
For instance, the decay parameter $\lambda$ is only considered if the optimizer $o \in \{ \text{Adam}, \text{ASGD} \}$ is ASGD.
Second, the optimizer influences the model's architecture. Specifically, it affects the set of possible values for the number of hidden layers and {number of units} in each included layer that is represented by a "requires" relationship, as it changes the possible bounds. Choices are represented by "alternative" features, as \textcolor{black}{the optimizer that is either ASGD  or Adam but not both, whereas the included or not layers are represented by OR features. {For the sake of illustration, we have no incompatibilities and therefore, no "excludes" relationship on this problem.}}

\noindent The feature model of Fig.~\ref{fig:MLP_FM} presents several limitations. First, continuous variables have non-discrete support that cannot be addressed by feature models. However, this remark can partially be addressed by a feature model extension called attribute feature models~\cite{abbas2016multi,streitferdt2003details}. 

Attribute feature models address a key limitation by explicitly encoding dependencies that are otherwise implicit. For example, they capture that including a hidden layer requires specifying how many units it may contain and they also clarify how the choice of optimizer constraints which architectures are valid, and enable more flexible, partially specified variable definitions by explicitly defining allowable value ranges (supports).
Attributes do not necessarily serve as variable supports, and features do not always correspond to optimization or modeling variables. For example, in the Feature Model (FM), "Adam" is considered a feature, whereas in the design space definition, it represents a level of the variable "Optimization". Similarly, "hidden layers" may be a feature in the FM but not a design variable in the design space. In the design space, the number of hidden layers is treated as a variable, while in the FM, this choice is governed by an "OR" relationship, meaning that determining the number of hidden layers requires counting how many have been selected. These distinctions highlight the need for the hierarchical model introduced in this paper. Unlike traditional feature models and design space graphs, this model is more directly aligned with design variables, making it a complementary approach that bridges the gap between existing representations.

\begin{figure}[!h]
\centering
    \scalebox{0.65}[0.70]{

\begin{tikzpicture}
\fontsize{12}{12}\selectfont
\begin{scope}[shift={(0, -0.25cm)}]

\node(opt)[label={Optimizer}, draw, ellipse] at (0,0)
{\begin{tabular}{c}
$o $ \\ 
$ \in \{ \text{ASGD}, \text{Adam} \}$
\end{tabular}} ;

\node(l1)[label={Decay}, xshift=-9cm, yshift=-2.5cm, at=(opt), draw, ellipse]
{\begin{tabular}{c}
$\alpha_1$ \\
$\in A_1(o)$
\end{tabular}};
\draw[->, dotted, thick] (opt.west)--(l1) node[midway, anchor=east, above, xshift=0cm, yshift=-1cm] {};

\node(l2)[label={Power update}, xshift=-6cm, yshift=-2.5cm, at=(opt), draw, ellipse]
{\begin{tabular}{c}
$\alpha_2$ \\
$\in A_2(o)$
\end{tabular}};
\draw[->, dotted, thick] (opt.west)--(l2) node[midway, anchor=east, above, xshift=0cm, yshift=-1cm] {};

\node(l3)[label={Average start}, xshift=-3cm, yshift=-2.5cm, at=(opt), draw, ellipse]
{\begin{tabular}{c}
$\alpha_3$ \\
$\in A_3(o)$
\end{tabular}};
\draw[->, dotted, thick] (opt.west)--(l3) node[midway, anchor=east, above, xshift=0cm, yshift=-1cm] {};

\draw [decorate, decoration = {calligraphic brace, amplitude=15pt}, yshift=0cm,] ([xshift=1cm]l3.south) --  ([xshift=-1cm]l1.south) node [black, midway, xshift=0cm, yshift=-0.75cm] {Hyperparameters associated ASGD};

\node(l4)[label={\# of layers}, xshift=0cm, yshift=-2.5cm, at=(opt), draw, ellipse]
{\begin{tabular}{c}
$l$ \\
$\in L(o)$
\end{tabular}};
\draw[->, dotted, thick] (opt.south)--(l4) node[midway, anchor=east, above, xshift=0cm, yshift=-1cm] {};

\node(l5)[label={Run. average 1}, xshift=3cm, yshift=-2.5cm, at=(opt), draw, ellipse]
{\begin{tabular}{c}
$\beta_1$ \\
$\in B_1(o)$
\end{tabular}};
\draw[->, dotted, thick] (opt.east)--(l5) node[midway, anchor=east, above, xshift=0cm, yshift=-1cm] {};

\node(l6)[label={Run. average 2}, xshift=6cm, yshift=-2.5cm, at=(opt), draw, ellipse]
{\begin{tabular}{c}
$\beta_2$ \\
$\in B_2(o)$
\end{tabular}};
\draw[->, dotted, thick] (opt.east)--(l6) node[midway, anchor=east, above, xshift=0cm, yshift=-1cm] {};

\node(l7)[label={Num. stability}, xshift=9cm, yshift=-2.5cm, at=(opt), draw, ellipse]
{\begin{tabular}{c}
$\beta_3$ \\
$\in B_3(o)$
\end{tabular}};
\draw[->, dotted, thick] (opt.east)--(l7) node[midway, anchor=east, above, xshift=0cm, yshift=-1cm] {};

\draw [decorate, decoration = {calligraphic brace, amplitude=15pt}, yshift=0cm,] ([xshift=1cm]l7.south) --  ([xshift=-1cm]l5.south) node [black, midway, xshift=0cm, yshift=-0.75cm] {Hyperparameters associated to Adam};


\node(u1)[label={Units 1st layer}, xshift=-7cm, yshift=-6.5cm, at=(opt), draw, ellipse]
{\begin{tabular}{c}
$u_1$ \\
$\in U_1(o,l)$
\end{tabular}};
\draw[->, dotted, thick] (l4.south)--(u1) node[midway, anchor=east, above, xshift=0cm, yshift=-1cm] {};

\node(u2)[label={Units 2nd layer}, xshift=-3cm, yshift=-6.5cm, at=(opt), draw, ellipse]
{\begin{tabular}{c}
$u_2$ \\
$\in U_2(o,l)$
\end{tabular}};
\draw[->, dotted, thick] (l4.south)--(u2) node[midway, anchor=east, above, xshift=0cm, yshift=-1cm] {};

%
\node(u3)[label={}, xshift=3cm, yshift=-6.5cm, at=(opt), , ellipse]
{\begin{tabular}{c}
{\LARGE \ldots}
\end{tabular}};
\draw[->, dotted, thick] (l4.south)--(u3) node[midway, anchor=east, above, xshift=0cm, yshift=-1cm] {};

\node(u4)[label={Units $l^{\text{max}}$-th layer}, xshift=7cm, yshift=-6.5cm, at=(opt), draw, ellipse]
{\begin{tabular}{c}
$u_{l^{\text{max}}}$ \\
$\in U_{l^{\text{max}}}(o,l)$
\end{tabular}};
\draw[->, dotted, thick] (l4.south)--(u4) node[midway, anchor=east, above, xshift=0cm, yshift=-1cm] {};

\node(r)[label={Learning rate}, xshift=-7.25cm, yshift=3cm, at=(l4), draw, ellipse]
{\begin{tabular}{c}
$r$ \\
$\in \ (0,1)$
\end{tabular}};

\node(a)[label={Activation function}, xshift=7.25cm, yshift=3cm, at=(l4), draw, ellipse]
{\begin{tabular}{c}
$a$ \\
$\in \ \{ \text{ReLU}, \text{Sig}, \text{Tanh} \}$
\end{tabular}};

\end{scope}
\end{tikzpicture}

    }
\caption{\textcolor{black}{Role graph $G$ for the MLP working example.}}
\label{fig:working_example_roles}
\end{figure}
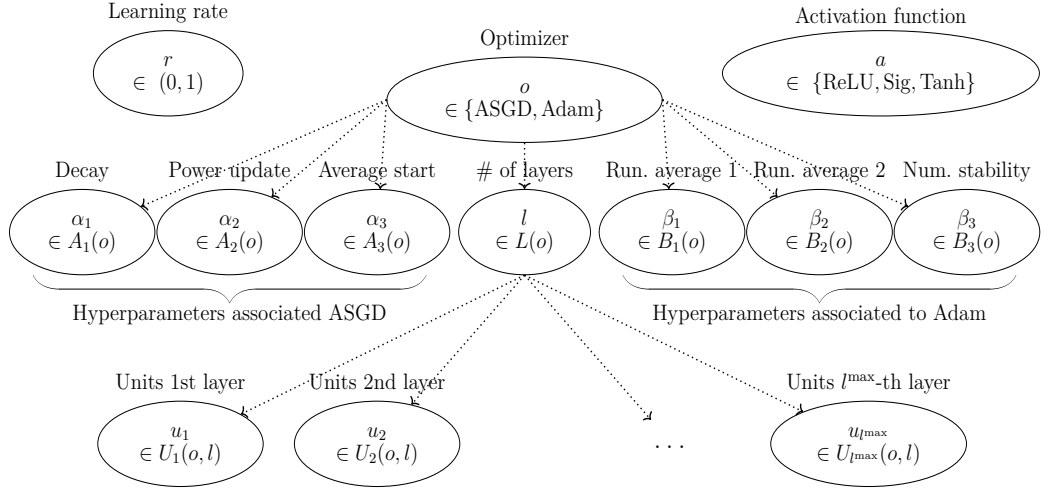

With the design space graph approach, the first modeling step is to build the role graph $G=(N,A)$, which is presented schematically in Fig.~\ref{fig:working_example_roles}.
Note that the index $i$ in $u_i$ indicates the $i$-th hidden layer.
For example, if $l=3$, there are three hidden layers, each with its own hyperparameter for the number of units, \textit{i.e.}, $u_1, u_2, u_3$.
Furthermore, the number of hidden layers $l \in L_o$ (influenced by the chosen optimizer) also determines the number of variables associated to the units $u_i$ in the hidden layers, creating a 2-level hierarchy.

The role graph $G$ in Fig.~\ref{fig:working_example_roles} models a great deal of information, including 1) the decree relations, represented by arcs, between the variables, 2) the role of each variable via the position of its corresponding node in the DAG, 3) the depth of each variable, and 4) the dependencies of the belonging sets of meta-decreed and decreed variables with their lower-depth variables. As such, it is more explicit than the feature model. \textcolor{black}{The FM has 4 first-level features, whereas our hierarchical model has 3 roots because the number of hidden layers that we allow depends from the choice of the optimizer.} Therefore, to solve these incompatibilities, the graph model proposes to choose first the optimizer and then the number of hidden layers.

\textcolor{black}{Then, within the design space graph approach, the next modeling step consists in formulating the problem—such as an optimization or analysis task—based on specific variables whose admissible bounds are explicitly defined for exploration. This step resulted in the construction of Table~\ref{tab:mlp}, which summarizes the problem description and all its variable roles, types, and bounds.}

\begin{table}[!ht]
\centering
 \caption{\textcolor{black}{Definition of the ``\texttt{MLP}'' optimization problem.}}
\footnotesize

\footnotesize
\begin{tabular*}{\linewidth}{llllrr}
 & Function/variable & {Type} &  {Role}  &Quantity & Range\\
\hline
\hline
Minimize & Loss function & {cont} & & 1 &\\
 & \multicolumn{2}{l}{\bf Total objectives} && {\bf 1} & \\
\hline
with respect to  & \mbox{Learning rate} & {cont} & neutral & 1 & $(0, 1)$ \\  
& \mbox{Decay} & {cont} & decreed & 1 & $(0, 1)$ \\ 
& \mbox{Power update} & {cont} & decreed & 1 & $(0, 1)$ \\  
& \mbox{Average start} & {cont} & decreed & 1 & $(10^{3}, 10^8)$ \\  
& \mbox{Run. average 1} & {cont} & decreed & 1 & $(0, 1)$ \\  
& \mbox{Run. average 2} & {cont} & decreed & 1 & $(0, 1)$ \\  
& \mbox{Num. stability} & {cont} & decreed & 1 & $(0, 1)$ \\  

 & \multicolumn{2}{l}{\bf Total  continuous variables} & & 7 & \\
 \cline{2-5}
  & \mbox{Activ. funct.} & {cat} & {neutral} & 3 levels & \{\text{ReLU},\text{Sigmoid},\text{Tanh}\} \\
 & \mbox{Optimizer} & {cat} & {meta} & 2 levels & \{\text{ASGD},\text{Adam}\} \\
  & \mbox{Number of layers} & {int} & {meta-decreed} & 1 & \{1,2,3\} \\
& \mbox{Neur. Units layer 1} & {int} & decreed & 1 & $\{25,30,\ldots,45 \}$ \\ & \mbox{Neur. Units layer 2} & {int} & decreed & 1 & $\{25,30,\ldots,45 \}$ \\ 
& \mbox{Neur. Units layer 3} & {int} & decreed & 1 & $\{25,30,\ldots,45 \}$ \\  
 
\hline
\end{tabular*}
\label{tab:mlp}
\end{table}

\subsection{Modeling a multidisciplinary aircraft system architecture}
\label{sec:ad}
    \textcolor{black}{
Designing complex engineering systems, such as aircraft, requires coordinating multiple interacting disciplines, including aerodynamics, structures, propulsion, and controls, where decisions in one area produce feedback that propagates through others and creates tightly coupled, often nonlinear behavior. To manage this complexity, eXtended Design Structure Matrix (XDSM) models offer a concise yet expressive way to represent modules, data flows, solver loops, and iteration order, making it easy to see which subsystems must be solved iteratively and which can be treated more loosely~\cite{Lambe2012}. These abstract system models serve as blueprints that can be translated into executable Multidisciplinary Design Analysis (MDA) and optimization simulations in frameworks such as OpenMDAO, where each discipline is implemented as a component, drivers coordinate solvers and optimizers, and system-level derivatives or surrogates can be propagated for efficient gradient-based design exploration~\cite{gray2019openmdao}. Together, XDSM and an MDAO improve traceability and reproducibility, guide solver and surrogate choices, and accelerate disciplined trade studies before committing to costly high-fidelity analyses.}

\textcolor{black}{
The \texttt{DRAGON} baseline configuration is an aircraft featuring two turboshafts, four generators, four propulsion buses with cross-feed, and forty fans. This configuration was selected for the initial study as it satisfies the safety criterion. However, it was not designed to optimize aircraft weight. The turboelectric propulsive chain being an important weight penalty, it is of particular interest to optimize the chain and particularly the number and type of each component, characterized by some mixed hierarchical values. 
In our workflow, the high-level XDSM representations are translated into executable MDA simulations, forming the computational core of the DRAGON co-design model. These simulations integrate aerodynamic, structural, propulsion, and electrical models within mission-analysis and sizing loops, ensuring that all estimated quantities reflect a physically consistent equilibrium achieved through multidisciplinary coupling and convergence towards consensus across disciplines.
The simulations rely on the open-source Future Aircraft Sizing Tool with Overall Aircraft Design (FAST-OAD)~\cite{David_2021}, enhanced with expert knowledge and recent extensions for hybrid-electric and distributed propulsion concepts. Built upon the OpenMDAO framework~\cite{gray2019openmdao}, FAST-OAD provides a modular structure where each discipline acts as a component, and where the data exchanges and solver couplings are explicitly defined. This setup enables flexible management of multidisciplinary interactions and clear control of convergence behavior.}

The consideration of the variables related to architecture was revised to take full advantage of mixed variables optimization. Three configurations with different numbers of electric components were considered, each with its own sizing rules. The default configuration is preserved for the analysis, and two new configurations, one with low distribution and the other with high distribution, were created and analyzed to establish the sizing rules. The number of motors was to remain an optimization variable as it is an important driver of the propulsive and aerodynamic efficiency of the aircraft, but the type of architecture constrains it. In this analysis, the number of motors could only be a multiple of 4, 8, or 12, depending on the selected architecture. The solution consisted of expanding the levels of the categorical variable representing the architecture and assigning a specific and valid number of motors to each level.

In a previous work, we optimized the Distributed fans Research Aircraft with electric Generators by ONERA (DRAGON), but without considering any hierarchy~\cite{SciTech_cat}, and an additional goal of this article is to show how our framework can be applied in practice.  
The definition of the turboshaft layout is given in~\tabref{tab:dragon_archi2c6} and the definition of the architecture variable is given in~\tabref{tab:dragon_archi1c6}\textcolor{black}{, where, for the sake of simplicity, we fixed the number of generators to be equal to the number of cores.}
\begin{table}[!h]
\centering
\vspace*{-0.3cm}

 \caption{Definition of the turboshaft layout variable and its 2 associated levels.}

\footnotesize

\begin{tabular*}{\linewidth}{cccccc}
\hline
  \textbf{Layout number} & position & y ratio & tail & VT aspect ratio & VT taper ratio\\
  \hline 
  \textbf{1} & under wing &0.25 & without T-tail& 1.8 & 0.3 \\
  \textbf{2} & behind & 0.34 & with T-tail& 1.2 & 0.85\\
 
\hline
\end{tabular*}
\label{tab:dragon_archi2c6}
\end{table}

\begin{table}[!h]
\centering
\vspace*{-0.3cm}

\small
 \caption{Definition of the architecture variable and its 17 associated levels.}
\begin{tabular*}{\linewidth}{l@{\hspace{10pt}}c@{\hspace{5.1pt}}c@{\hspace{5.1pt}}c@{\hspace{5.1pt}}c@{\hspace{5.1pt}}c@{\hspace{5.1pt}}c@{\hspace{5.1pt}}c@{\hspace{5.1pt}}c@{\hspace{5.1pt}}c@{\hspace{5.1pt}}c@{\hspace{5.1pt}}c@{\hspace{5.1pt}}c@{\hspace{5.1pt}}c@{\hspace{5.1pt}}c@{\hspace{5.1pt}}c@{\hspace{5.1pt}}c@{\hspace{5.1pt}}c@{}}
\hline
\textbf{Architecture number} & \textbf{1} & \textbf{2} & \textbf{3} & \textbf{4} & \textbf{5} & \textbf{6} & \textbf{7} & \textbf{8} & \textbf{9} & \textbf{10} & \textbf{11} & \textbf{12} & \textbf{13} & \textbf{14} & \textbf{15} & \textbf{16} & \textbf{17} \\
\hline 
\textbf{\textcolor{black}{Number of cores}} & 2 & 2 & 2 & 2 & 2 & 2 & 2 & 2 & 2 & 4 & 4 & 4 & 4 & 4 & 4 & 4 & 4 \\
\textbf{Number of generator} & 2 & 2 & 2 & 2 & 2 & 2 & 2 & 2 & 2 & 4 & 4 & 4 & 4 & 4 & 4 & 4 & 4 \\
\textbf{\textcolor{black}{Number of motors}} & 8 & 12 & 16 & 20 & 24 & 28 & 32 & 36 & 40 & 8 & 16 & 24 & 32 & 40 & 12 & 24 & 36 \\
\hline
\end{tabular*}
\label{tab:dragon_archi1c6}
\end{table}

\textcolor{black}{
At the time, the propulsive architecture had to be encoded as a single categorical variable with 17 possible levels (Table~\ref{tab:dragon_archi1c6}).
Notwithstanding, for modeling electric architectures, it is more efficient to represent the architectural choices using two integer variables instead of one categorical variable. 
However, taking this approach expands the range of potential architectures beyond the initial 17 configurations. Yet, there are two important constraints to consider for these possible setups.
The first constraint relates to the electrical connections between components: ensuring a certified electric architecture is crucial, and figuring out how to connect, for example, 8 motors to 6 generators is not straightforward.
The second constraint is connected to the distributed propulsion system, especially the numerous propellers. Managing this system involves addressing a substantial number of potential failures in the electro-mechanical architecture, as for both stability and redundancy, not all electric connections are allowed.
Using the hierarchical framework introduced in this paper, this single categorical choice can be decomposed into the two hierarchical integer variables that encode the propulsive chain. Not only do we have a better representation of the architecture system, but, as a side effect, in the continuous relaxation setup, this reduces the number of design variables associated with the propulsive architecture from 17 to 3, thereby simplifying and accelerating continuous relaxation based optimization for the overall problem.
}
\textcolor{black}{The problem defined in Table~\ref{tab:dragonc6} targets an expensive-to-evaluate, high-fidelity blackbox simulation. The objective is to minimize the aircraft fuel mass over the mission—a key driver of both environmental impact and operating cost—subject to five expensive-to-evaluate system-level constraints. These constraints are outputs of the simulator (\textit{i.e.}, they result from the simulator's internal computations and couplings given the input variables) and therefore differ from algebraic or structural constraints imposed directly on the inputs.}

\begin{table}[!h]
\centering
 \caption{Definition of the \texttt{DRAGON} optimization problem.}
\footnotesize

\footnotesize
\begin{tabular*}{\linewidth}{llllrr}
 & Function/variable & \textcolor{black}{Type} &  \textcolor{black}{Role}  &Quantity & Range\\
\hline
\hline
Min. & Fuel mass & \textcolor{black}{cont} & & 1 &\\
 & \multicolumn{2}{l}{\bf Total objectives} & & {\bf 1} & \\
\hline
w.r.t & \mbox{Fan operating pressure ratio} & \textcolor{black}{cont}& \textcolor{black}{neutral} & 1 & $\left[1.05, 1.3\right]$ \\  
     & \mbox{Wing aspect ratio} & \textcolor{black}{cont} & \textcolor{black}{neutral} & 1 &    $\left[8, 12\right]$ \\
    & \mbox{Angle for swept wing} & \textcolor{black}{cont} & \textcolor{black}{neutral} & 1 & $\left[15, 40\right]$  ($^\circ$) \\
     & \mbox{Wing taper ratio} & \textcolor{black}{cont} & \textcolor{black}{neutral}  & 1 &    $\left[0.2, 0.5\right]$ \\
     & \mbox{HT aspect ratio} & \textcolor{black}{cont} & \textcolor{black}{neutral} & 1 &    $\left[3, 6\right]$ \\
    & \mbox{Angle for swept HT} & \textcolor{black}{cont} & \textcolor{black}{neutral} & 1 & $\left[20, 40\right]$  ($^\circ$) \\
     & \mbox{HT taper ratio} & \textcolor{black}{cont} & \textcolor{black}{neutral}  & 1 &    $\left[0.3, 0.5\right]$ \\
 & \mbox{TOFL for sizing}  & \textcolor{black}{cont} & \textcolor{black}{neutral}  &1 & $\left[1800., 2500.\right]$ ($m$) \\
 & \mbox{Top of climb vertical speed for sizing} & \textcolor{black}{cont} & \textcolor{black}{neutral} & 1 & $\left[300., 800.\right]$($ft/min$) \\
 & \mbox{Start of climb slope angle} & \textcolor{black}{cont} & \textcolor{black}{neutral} & 1 & $\left[0.075., 0.15.\right]$($rad$) \\

 & \multicolumn{2}{l}{ \bf Total  continuous variables} & & 10 & \\
 \cline{2-5}
 & \mbox{Turboshaft layout} & \textcolor{black}{cat} & \textcolor{black}{neutral} & 2 levels & \{1,2\} \\
\hdashline[0.5pt/5pt]

\hdashline[0.5pt/5pt]
& \mbox{{Number of cores}} & \textcolor{black}{int} & \textcolor{black}{meta} & 1 & \{2,4,6\} \\
& \mbox{{Number of motors}} & \textcolor{black}{int} & \textcolor{black}{decreed} & 1 & \{8,12,16,20,\ldots,40\} \\
\hdashline[0.5pt/5pt]
 \cline{2-5}  
s.t. & Wing span \textless  \ 36   ($m$)  & \textcolor{black}{cont} & & 1 \\
 & TOFL \textless  \ 2200 ($m$) & \textcolor{black}{cont} & &  1 \\
 & Wing trailing edge  & \textcolor{black}{cont} & & 1 \\
 & \multicolumn{3}{l}{ $\ $ occupied by fans  \textless  \ 14.4 ($m$) } \\
 & Climb duration \textless  \ 1740 ($s $) & \textcolor{black}{cont} && 1 \\
 & Top of climb slope \textgreater \ 0.0108 ($rad$) & \textcolor{black}{cont} && 1 \\
 & \multicolumn{2}{l}{\textbf{Total  constraints}} & & {\textbf{5}} & \\
\hline
\end{tabular*}
\label{tab:dragonc6}
\end{table}

\subsection{The graph-structured design space: implementation and usage}
\label{sec:gsds}

Hereinafter, we extend the framework from~\cite{saves2023smt} by incorporating the key features reviewed in Sect.~\ref{sec:intro_literature_review}. Our implementation builds on SMT 2.0’s design‐space tools~\cite{saves2023smt,SMT2019}, the ConfigSpace library~\cite{SMAC3}, and the DSG available in the software \texttt{adsg-core}~\cite{bussemaker2024ADSG,bussemaker2022adore}, making it practical for real‐world use\footnote{\url{https://adsg-core.readthedocs.io/}}.
Table~\ref{tab:comparisonsoft} outlines the main features of the Python modeling software that can handle hierarchical and mixed variables within SMT, chosen within the open-source software literature based on its mixed variable handling capacities as detailed in~\cite[Table 1]{saves2023smt} where several are compared for such modeling task. The default option in SMT is a basic DSG that is limited to one level of hierarchy and is not able to handle complex \textcolor{black}{relationships} such as incompatibility constraints, even if hierarchical. The ConfigSpace design space may be slow and is not based on the graph structure. Overall, the best option for architecture modeling is the DSG of the \texttt{adsg-core} toolbox that is fast and gives explicit visualizations of the DSG in DOT language, \textcolor{black}{draw.io or GML}~\cite{ellson2002graphviz} and our main software contribution has been combining DSG to work within SMT to define hierarchical surrogate models.

\begin{table}[!h]
\centering
\resizebox{\columnwidth}{!}{%
\begin{tabular}{l c c c c c c c}
\hline
\textbf{Package} 
& \textbf{License} 
& \makecell[c]{\texttt{SMT} \\ interface name}  
& \makecell[c]{Mixed \\ hier.\ var.}   
& \makecell[c]{Nested \\ hier.} 
& \makecell[c]{Incomp. \\ (excl.)} 
& \makecell[c]{Graph \\ struct.} 
& \makecell[c]{Explicit \\ \& visu.} \\
\hline
\textbf{SMT}~\cite{saves2023smt} 
  & BSD  
  & \texttt{DesignSpace} 
  & \checkmark  
  & --  
  & --  
  & \checkmark  
  & --  \\

\textbf{ConfigSpace}~\cite{SMAC3} 
  & BSD  
  & \texttt{ConfigSpaceDesignSpaceImpl} 
  & \checkmark 
  & \checkmark 
  & \checkmark  
  & --  
  & --  \\

\textbf{adsg-core}~\cite{bussemaker2024ADSG} 
  & MIT  
  & \texttt{AdsgDesignSpaceImpl} 
  & \checkmark 
  & \checkmark 
  & \checkmark  
  & \checkmark  
  & \checkmark  \\
\hline
\end{tabular}
}
\caption{Comparison of software packages for hierarchical and mixed design‐space definition.}
\label{tab:comparisonsoft}
\end{table}

We can automatically define a DSG from the variables specifications\footnote{\url{https://github.com/SMTorg/smt-design-space-ext/}}.

The DSG models hierarchical choices using selection and connection choice nodes, that select between mutually exclusive option nodes and connect sets of source nodes to target nodes, respectively~\cite{bussemaker2024ADSG}.
Nodes are subject to selection and only obtain meaning in the context of an optimization problem.
Its original context was function-based modeling of system architecture design spaces, called the Architecture Design Space Graph (ADSG), where various types of nodes, such as functions, components, and ports, were defined~\cite{BuCiNa2020}.

\subsection{Application of the graph-structured design space to the MLP problem}
\label{sec:mlpapp}
For illustrating the method, let  \texttt{design\_space} be the SMT 2.0 \texttt{AdsgDesignSpaceImpl} defined in Appendix~\ref{app:pseudo-code} (Fig.\ref{fig:ds_nn}) for the \textcolor{black}{MLP problem of Sect.~\ref{sec:working_example_MLP}}.
In a DSG, both neutral and purely meta variables serve as graph roots, also referred to as "permanent nodes" or "start nodes", while decreed and meta-decreed variables function as "conditional nodes". A choice that depends on a meta or meta-decreed variable is represented by a choice node (in blue) and is connected to the variable name and its possible choices through derivation edges.
Future work may include automated formal verification of the graph structure.
However, some verification steps are already handled within the DSG. For instance, the DSG is considered infeasible if an incompatibility edge is defined between two permanent nodes, ensuring that conflicting configurations are detected early. \

To do this, the \textit{encoder} maps architectural choices, such as selections and connections in the graph, into a numerical design vector $\bold{x}$ that optimizers can handle.
Therefore, selection and connection choice encoders enable a full enumeration of valid design vectors and ensure any source-to-target connection problem can be encoded such that optimization algorithms can effectively search the design space. The encoder plays a crucial role in mapping selection choices from the hierarchical space to discrete variables in the input vectors, ensuring correct associations with input points for tasks such as sampling.  
Currently, two encoding methods are implemented. The \textit{complete encoder} performs an exhaustive exploration of the design space, identifying all valid configurations and providing accurate imputation ratios. In contrast, the \textit{fast encoder} offers a more efficient but less comprehensive approach by directly mapping selection choices to design variables. While the fast encoder significantly reduces computation time, it may overlook some valid configurations, making it more suitable for large design spaces with computational constraints.  
\textcolor{black}{The complete encoder implements a constrained enumeration: it derives local connection choice formulations from the DSG, builds an influence matrix to prune infeasible partial assignments, and enumerates the set of valid connection matrices to produce a full mapping that permits exact decoding or nearest-valid lookup.  
By contrast, the fast encoder avoids global enumeration and decodes greedily, trading completeness for much lower memory and runtime requirements and serving as the practical fallback when exhaustive enumeration is infeasible. More details are given in~\cite[Appendix B]{bussemaker2025system}.
}

\begin{figure}[!h]
    \centering
     \includegraphics[width=\textwidth]{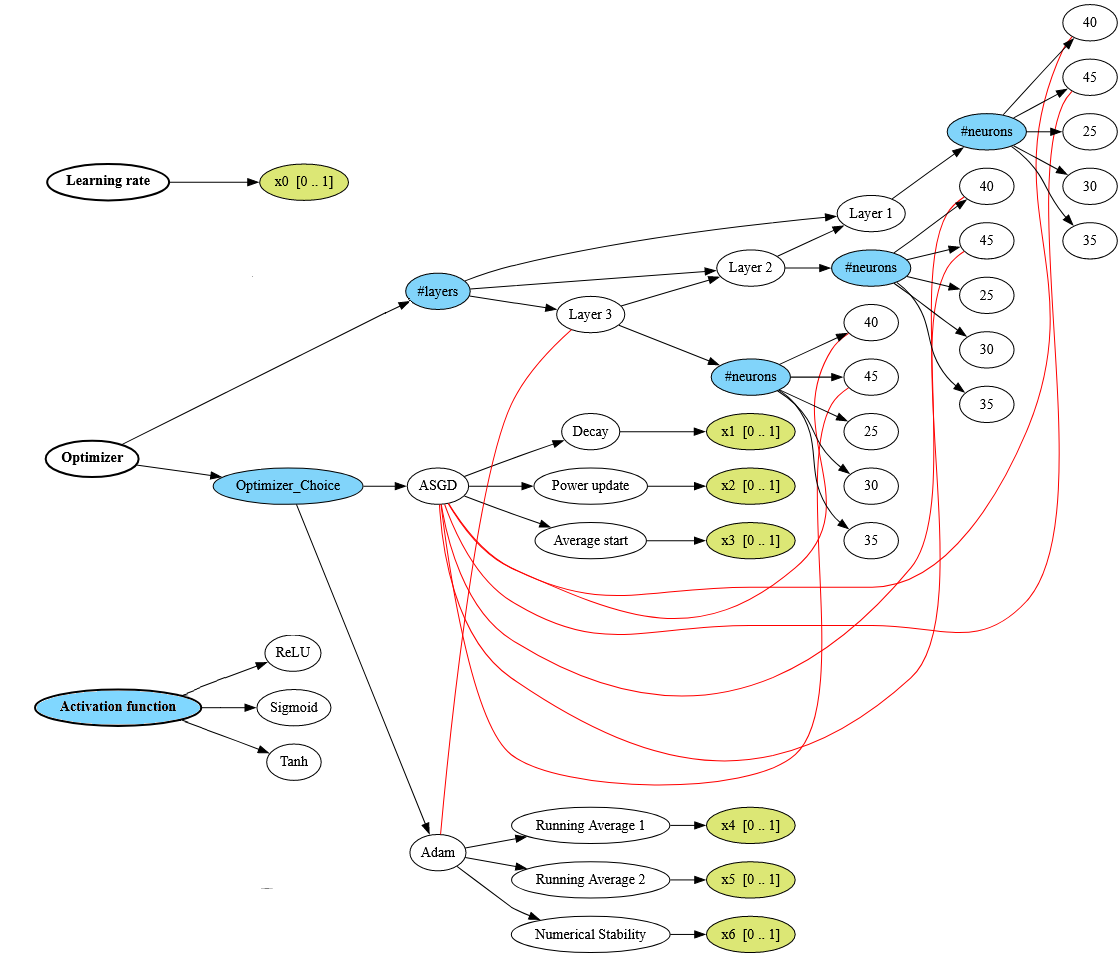}
    \caption{DSG definition for the neural network problem.}
    \label{fig:adsg_nn}
 \end{figure}

To construct the DSG shown in Fig.~\ref{fig:adsg_nn}, we only need to define the variables, levels, and relationships. Then, the DSG is processed to have a visual graph with the nodes in $N_{\text{var}}$ in blue if categorical and white otherwise, the nodes in $N_{\text{levels}}$ in white and those in $N_{\text{bnd}}$ in {straw color}. The edges in $N_{\text{opposed}} $ are colored in red.
The DSG also automatically processes the graph to generate outputs, as presented in Table~\ref{fig:adsg_nn_out}.

\begin{table}[!h]

\resizebox{0.99\columnwidth}{!}{%
\begin{tabular}{lcccccccccccc}
\hline
\textbf{Type} & \textbf{n\_valid} & \textbf{n\_declared} & \textbf{n\_discrete} & \textbf{n\_dim\_cont} & \textbf{n\_dim\_cont\_mean} & \textbf{imp\_ratio} &  \textbf{encoder} \\
\hline
option-decisions     & 207 & 1350 & 6  & 0  & 0.0  & 6.521739  &  complete \\
additional-dvs       & 207 & 0    & 0  & 7  & 4.0 &  1.750000    &           \\
total-design-space   & 207 & 2250 & 6  & 7  & 4.0  & 10.869565   &  complete \\
\hline
\end{tabular}
}
\caption{Processed DSG on the neural network problem.}
\label{fig:adsg_nn_out}
\end{table}

\textcolor{black}{This analysis provides useful insights, including the {imputation ratio} (\texttt{imp\_ratio}), which measures the ratio between the total number of declared categorical and integer possibilities (\texttt{n\_declared}) and the number of feasible instances  (\texttt{n\_valid}). In this case, 2250 possibilities are declared, but only 207 are valid, meaning that approximately 1 in 10.9 instances is feasible. The table also displays the total number of variables, which consists of 13 variables: 6 discrete (\texttt{n\_discrete}) and 7 continuous (\texttt{n\_dim\_cont}). The {continuous imputation ratio} is 1.75, indicating that only 4 out of the 7 continuous variables are included (\texttt{n\_dim\_cont\_mean}). Specifically, the included variables correspond to either the combination of "decay", "Power update", and "Average start" or the combination of "Running average 1", "Running average 2", and "Numerical Stability".    
The \texttt{encoder} column in Table~\ref{fig:adsg_nn_out} indicates use of the complete encoder; exhaustive enumeration therefore ran efficiently and found all valid design vectors among the 2250 possible combinations. } 

\begin{figure}[H]
\centering
\includegraphics[width=0.75\textwidth]{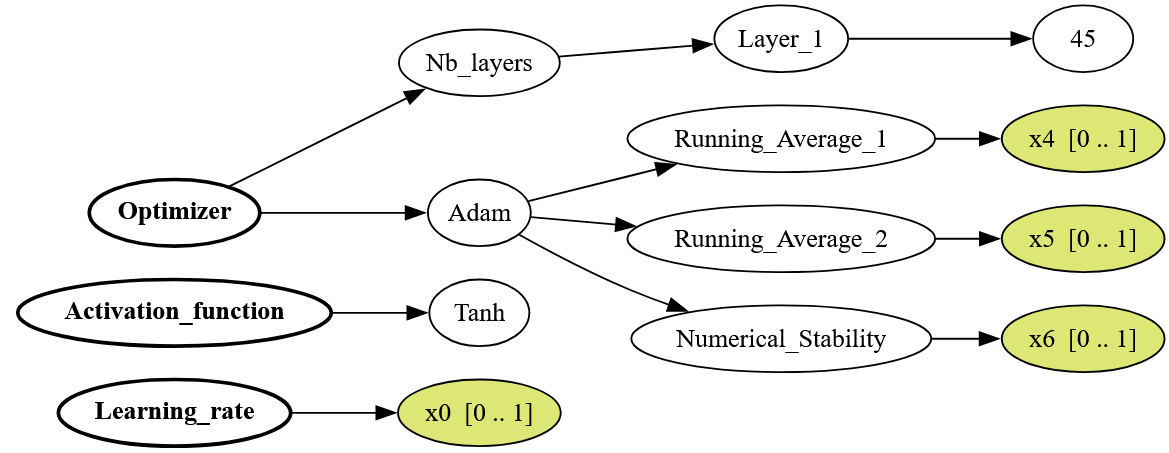}

\caption{\texttt{AdsgDesignSpaceImpl} sampling.}
\label{fig:ads_samp}
\end{figure}

Thanks to the DSG, we can make explicit the definition of the hierarchical model by giving the model directly and generating a valid configuration as in Fig.~\ref{fig:ads_samp}. This new class upgrades and improves on the prior \texttt{ConfigSpaceDesignSpaceImpl} class that did not take into consideration hierarchical spaces. The two can be compared on processing tasks as described 
 in~\tabref{tab:comparison}.   
The most important task consists of correcting a non-valid input point and returning both a corrected input and its corresponding included variable within the extended space (called active). This task of projecting an input from the extended domain to the hierarchical space is really important since the hierarchical distance is computed on the extended domain and then projected as this projection is an isomorphism~\cite[Theorem 2]{halle2024graph}. On this task, we observe, \textcolor{black}{in}~\tabref{tab:comparison} an improvement by $76$ \% with the \texttt{AdsgDesignSpaceImpl} that relies on adsg-core implementation compared to our legacy \texttt{ConfigSpaceDesignSpaceImpl} method relying on ConfigSpace.  
Another processing task is enumeration of all discrete possibilities, and in~\tabref{tab:comparison}, we obtained a 36\% speed up with the \texttt{AdsgDesignSpaceImpl} class for such a task. To finish with, one could want to generate a design of experiments that respects the graph-structure domain.
On this task, we obtain a 15\% improvement with our class, but to achieve this performance, we first generated one random point for each of the 207 possible discrete configurations and then subsampled 100 points from that set: an approach also adopted by SBArchOpt~\cite{bussemaker2023sbarchopt}. However, one might alternatively choose to sequentially sample additional points. While it is straightforward to parallelize the sampling of 100 points, doing so sequentially is approximately 15 times slower than the subsampling approach.
\begin{table}[!h]
\Large
\resizebox{0.99\columnwidth}{!}{%
\begin{tabular}{lccc}
\hline
\textbf{Class name} & $\begin{matrix}
\textbf{Correct and compute activeness } \\
\textbf{ of 1000 invalid points} \\
\end{matrix}$
& $\begin{matrix}
\textbf{Generate 1 point by} \\
\textbf{ discrete possibility (207 points)} \\
\end{matrix}$
 & \textbf{Generate 100 valid points} \\
\hline
\texttt{ConfigSpaceDesignSpaceImpl} &  2.2300s &  0.0081s & 0.0054s \\
\texttt{AdsgDesignSpaceImpl} &   0.5356s (-76\%)  &  0.0053s (-36\%) & 0.0046s (-15\%)\\
\hline
\end{tabular}
}
\caption{Comparison of processing duration for several tasks with \texttt{ConfigSpaceDesignSpaceImpl} and \texttt{AdsgDesignSpaceImpl}.}
\label{tab:comparison}
\end{table}

\textcolor{black}{In the next section, we will use the graph-structured DSG of SMT, which is based on the underlying DSG of ADSG, to create surrogate Gaussian process models. These surrogate models, therefore, support the modeling of hierarchical problems while preserving their structure. They are then applied to surrogate-based Bayesian optimization.}

\subsection{Bayesian optimization: implementation and usage}
\label{sec:bo}
Bayesian optimization has emerged as a highly effective technique for optimizing blackbox functions, particularly when these functions are expensive to evaluate, non-convex, or lack gradient information~\cite{frazierTutorialBayesianOptimization2018}. Its growing popularity in various fields, from machine learning hyperparameter tuning (HPO) to engineering design, is largely due to its ability to achieve near-optimal solutions with a minimal number of evaluations~\cite{cho2020basic}.
At its core, BO builds a probabilistic surrogate model, typically a GP, to approximate the objective function that is being optimized. The surrogate model provides not only predictions of the function's value at any given point in the search space but also an estimate of the uncertainty associated with these predictions. This uncertainty quantification method is the foundation of BO ability to balance exploration (sampling areas of the search space with high uncertainty) and exploitation (focusing on areas predicted to yield high objective values).
The BO process typically starts with a Design of Experiments (DoE) to obtain initial evaluations of the objective function. These evaluations are used to fit the GP model, which is then iteratively updated as new function evaluations are conducted. At each iteration, the next point to evaluate is chosen by maximizing an acquisition function to guide the search towards the optimum as fast as possible.

GP are well-suited for Bayesian optimization because they can flexibly represent unknown objective functions via a mean function and correlation kernel. The mean gives a prior estimate while the kernel encodes correlations across the search space, allowing uncertainty quantification and efficient predictions. In this work, we adopt the kernel described in Sect.~\ref{sec:GP-BO}, which balances smoothness assumptions with adaptability to mixed continuous, categorical, and hierarchical inputs.
Once the surrogate is set up using SMT’s \texttt{DesignSpace} class, defining a design of experiments and instantiating corresponding GP models—complete with graph-structured kernels—becomes a single, streamlined step.  
The acquisition function is a crucial component of BO, responsible for determining the next point to evaluate. Common acquisition functions include Expected Improvement (EI), Probability of Improvement (PI), Upper Confidence Bound (UCB) or Watson and Barnes $2^{\text{nd}}$ criterion scaled (WB2s)~\cite{Jones2001JOGO, ShahriariTakingHumanOut2016,bartoli:hal-02149236}.
In this work, we will use WB2s that are more smoothed and less multimodal than the other classical criteria, that is tractable and easy to optimize with a local optimizer such as COBYLA~\cite{COBYLA} and the optimizer SEGOMOE~\cite{bartoli:hal-02149236}.

In many real‐world optimization tasks, the design variables span continuous, integer, and categorical types, so standard continuous Bayesian optimization has been extended for them. Categorical inputs use specialized kernels, and, for comparison with our method, we use, from the simplest to the most complex method: Gower Distance (GD), Continuous Relaxation (CR), and Homoscedastic Hypersphere (HH) to capture categorical correlations, while integer variables employ ordinal‐aware, distance‐based kernels. More details on these methods are presented in~\cite{Mixed_Paul} as well as their unification in both modeling and software terms.
High dimensionality compounds the challenge, since modeling accuracy typically requires exponentially more samples. To mitigate this “curse of dimensionality”, dimensionality reduction methods such as Partial Least Squares (PLS) are integrated into the Gaussian‐process surrogate. In particular, Kriging with PLS (KPLS) has recently been generalized to mixed‐variable and even hierarchical spaces, enabling efficient BO over complex, heterogeneous domains~\cite{Mixed_Paul_PLS}. \textcolor{black}{However, the baseline surrogate model presented in this work, without any dimension reduction is limited to around one hundred variables, which may be sufficient in most applications, while the Architecture design space graph has been applied for up to 144 million configurations.}

\subsection{Application of the Bayesian optimization approach to the aircraft design problem}
\label{sec:boapp}
Once we have defined the surrogate model based on the \texttt{DesignSpace} class in SMT, we can directly define both DoE and surrogate models such as GP models while accounting for the hierarchical model. In~\cite{saves2023smt}, a GP model has been built for the hierarchical neural network problem and promising results have been obtained, and in~\cite{saves24}, the GP has been extended to handle decreed continuous variables and meta-decreed variables. 

To validate our method, we compare the 8 methods described in~\tabref{SMO_tab:dragon_meth} on the optimization of the \texttt{DRAGON} aircraft concept, with 10 points for the initial DoE. The first six are categorical kernels with different structures introduced and detailed in~\cite{Mixed_Paul}. The method \texttt{HIER} is the one based on the graph-structure design space and the last \texttt{NSGA-II} is an evolutionary algorithm posed as a baseline~\cite{DePrAgMe2002}. \textcolor{black}{We also tested dimension reduction, as such, several methods employ the PLS (Partial Least Squares) dimension reduction techniques~\cite{Mixed_Paul_PLS}.
These reduced-order models are given a target dimension that is the reduced size of their corresponding embeddings; for instance, \texttt{HH with PLS 3D} (or \texttt{HH\_3} for short) is the \texttt{HH} method reduced to 3 dimensions.
Even with these dimension reduction techniques, the matrix based methods are too costly for such problems: \texttt{HH} takes 320h, \texttt{HH with PLS 3D} takes 102h and \texttt{HH with PLS 12D} takes 258h.}
\begin{table}[!h]
\centering
 \caption{The various kernels compared for optimizing \texttt{DRAGON}. }
\small
\begin{tabular*}{\linewidth}{c@{\hspace{6pt}}c@{\hspace{4pt}}c@{\hspace{4pt}}c@{\hspace{4pt}}c@{\hspace{4pt}}c@{}}
\hline
    \textbf{Name} & \# of cat. params & \# of cont. params & Tot. \# of params  & Comp. time \\
    \hline 
    \texttt{GD} & 2 & 10 & 12 & 36h\\
     \hdashline
    \texttt{CR} & 19 & 10 & 29  & 62h \\
    \texttt{CR with PLS 3D} & Not applicable & Not applicable & 3 & 14h \\ 
    \hdashline
    \texttt{HH} & 137 & 10 & 147 & 320h \\
    \texttt{HH with PLS 3D} & 2 & 1 & 3 & 102h \\
    \texttt{HH with PLS 12D} & 2 & 10 & 12 & 258h \\
    \hdashline
    \texttt{NSGA-II} & Not applicable & Not applicable & Not applicable & 16h  \\
    \hdashline
    \texttt{HIER} & 1 & 12 & 13 & 40h \\
    \hline
\end{tabular*}
\label{SMO_tab:dragon_meth}
\end{table}   

\begin{figure}[H]
\subfloat[Convergence curves.\label{DRAGONcurves}]{
\includegraphics[clip=true,height=5.5cm, width=0.52\textwidth]{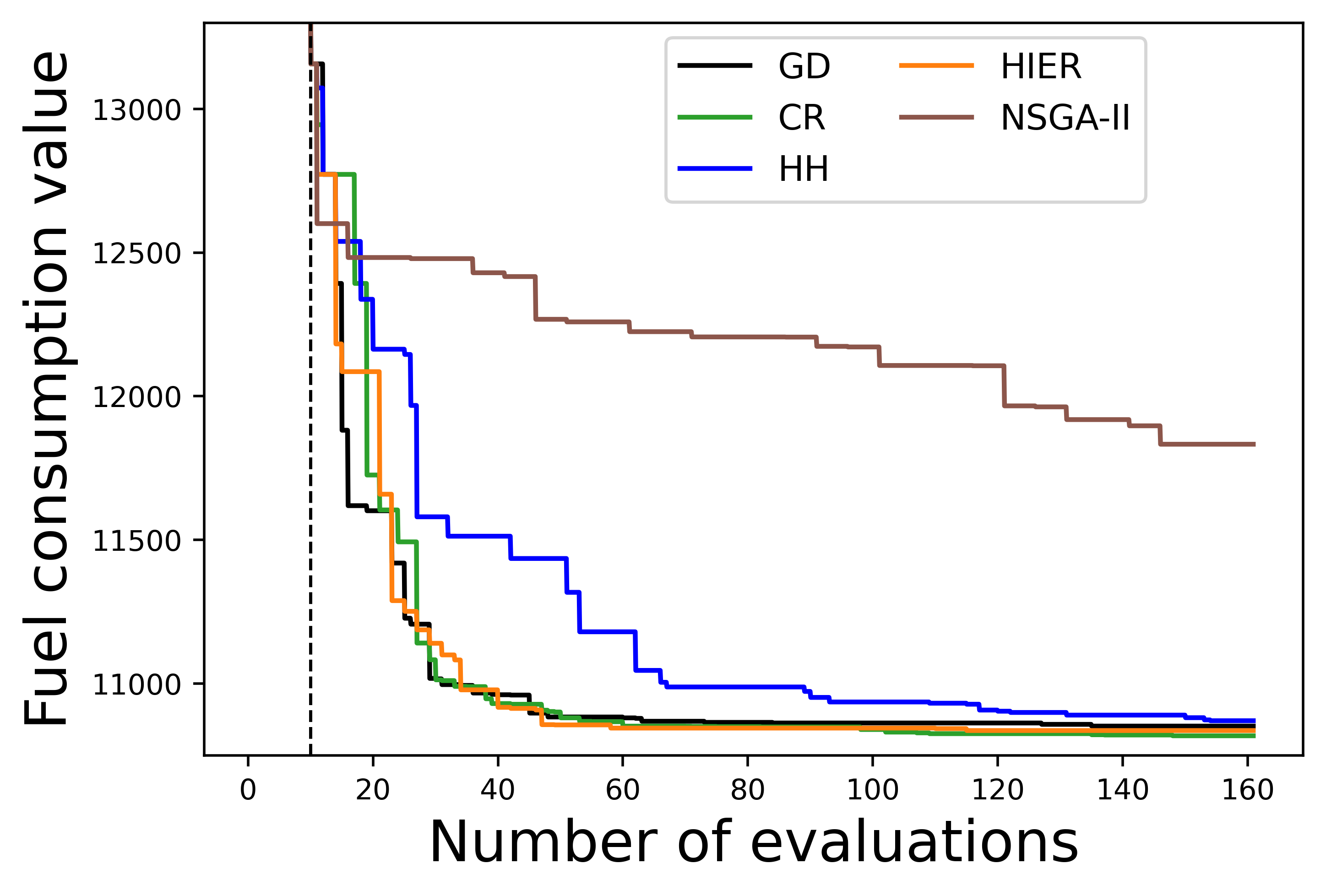}
}
\subfloat[Boxplots. \label{minima_DRAGON}]{
\includegraphics[clip=true,height=5.5cm, width=0.48\textwidth]{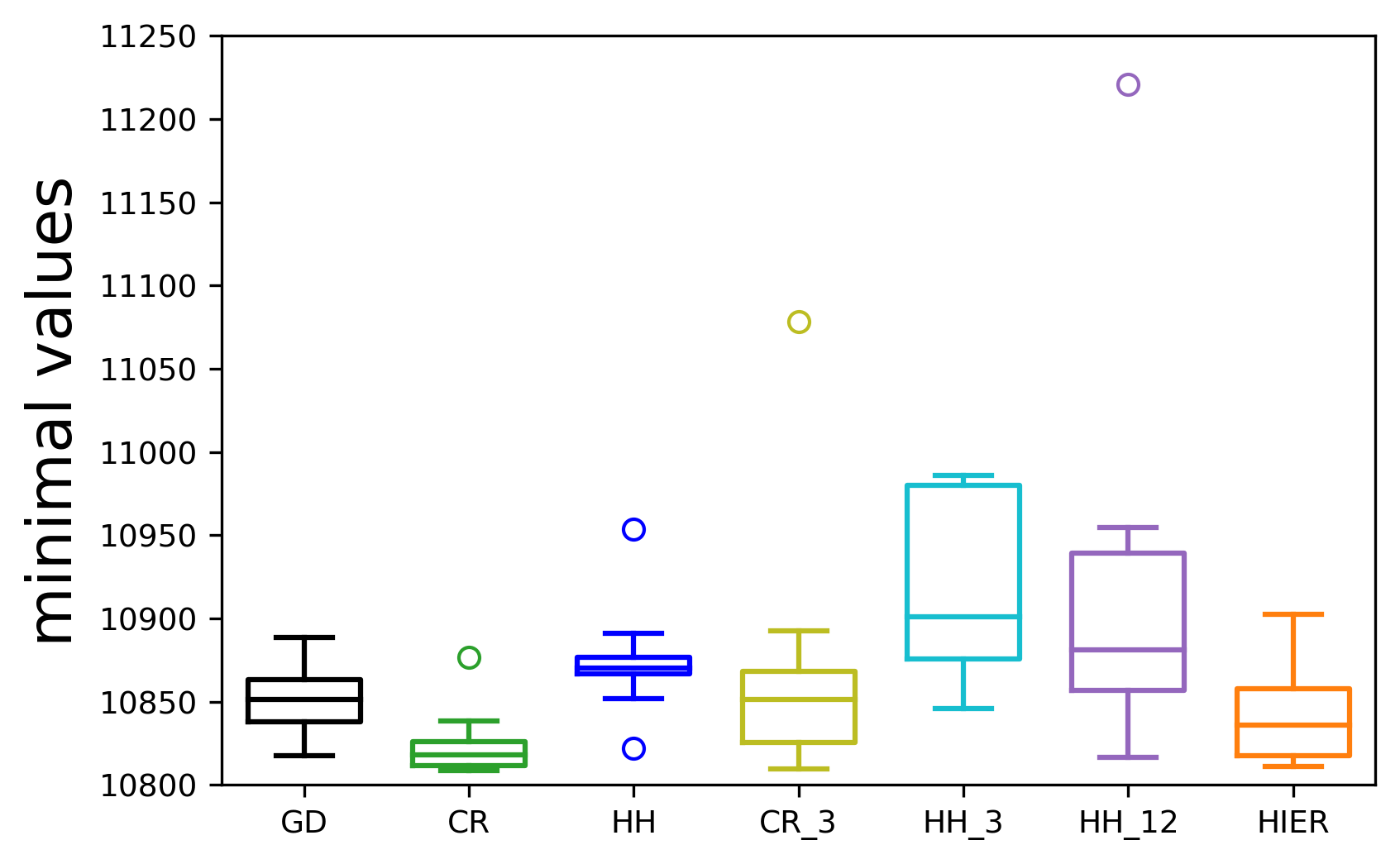}
}
\caption{\texttt{DRAGON} optimization results using a DoE of 10 points over 10 runs. The boxplots are generated, after 150 iterations, using the 10 best points.}
\label{DRAGON}
\end{figure}

Figure~\ref{DRAGON} shows promising results with the \textcolor{black}{hierarchical} domain. Namely, it is a bit more costly than \texttt{GD} for a better result and in terms of convergence behavior \texttt{GD}, \texttt{CR}, and \texttt{HIER} are highly similar. Note that \texttt{CR with PLS 3D} is faster than \texttt{HIER} for little degraded performance and that \texttt{CR} is the best in terms of convergence but is slightly more costly. In the end, \texttt{HIER} is a really good trade-off for BO.

\textcolor{black}{
The \texttt{DRAGON} case study illustrates the benefit of using hierarchical surrogate modeling within a Bayesian optimization framework for complex aircraft design problems. The hierarchical kernel (\texttt{HIER}) achieved a strong balance between accuracy and computational efficiency: while slightly more expensive than \texttt{GD}, it consistently provided better optima and convergence stability across all runs for only one parameter more (see Fig.~\ref{DRAGON}). In practical terms, the optimization led to a 439~kg reduction in fuel consumption for a single mission compared to the reference configuration, reaching 10{,}809~kg against 11{,}248~kg in the baseline configuration of the first studies~\cite{schmollgruber2}. The optimal configuration (option~10) features eight engines, four of which are hybrid turbo-generators placed at the rear, leveraging a favorable moment arm between the wing and horizontal tail. This design improves overall energy efficiency while maintaining feasible aerodynamic and mission performance. Although structural effects of high sweep and aspect ratio were simplified, the resulting architecture remains close to the reference optimum (10{,}816~kg) obtained in previous optimization work~\cite{saves2023smt}, confirming that the hierarchical surrogate and kernel-based exploration effectively capture trade-offs between structural layout, powertrain topology, and mission-level performance.
}

\section{Conclusion and Perspectives}
\label{sec:conclu}
This work introduced a novel operations research framework designed to define graph-structured design spaces, with application to surrogate modeling and optimization. We built upon previous works, extending their definitions to encompass any DAG, and we enhanced the flexibility and scalability of the framework. Specifically, this work extended the concept of meta and decreed variables presented in earlier research~\cite{audet2022general,saves24}, bridging the gap with design space graphs. 
Afterward, we demonstrate the improved framework capabilities and we proposed an industrial application for sizing and optimizing complex system architectures~\cite{BuCiNa2020,bussemaker2024system}. \textcolor{black}{Our graph-structured framework provides a flexible representation for mixed-variable and variable-size input feature spaces. It integrates naturally with surrogate-based optimization and is primarily intended for pointwise comparison rather than full graph-isomorphism tasks. However, it can become computationally burdensome for combinatorially large configuration spaces or when similarity depends on explicit topology matching. In such cases, we recommend combining the framework with scalable approximations or using specialized graph-kernel methods as appropriate~\cite{perez2024gaussian}.}

A key contribution of this research is the implementation of the extended framework within the open-source Surrogate Modeling Toolbox software~\cite{saves2023smt}.
This open-source toolbox is collaborative and is an easy to use and well documented platform for users and developers. 
Notably, SMT remains the only open-source toolbox capable of constructing both hierarchical and mixed surrogate models, and it has demonstrated its value in numerous practical applications, such as the optimization of a green aircraft design.
In short, this paper lays the groundwork for further developments involving hierarchical models or data. By improving both the theoretical foundation and practical implementation of this framework, the potential applications of this work are vast, ranging from industrial design optimization to more complex systems requiring advanced modeling and optimization techniques.

Several areas of future research and development have been identified. For instance, the new framework still lacks automated formal verification for graph structures. Similar to approaches used in feature models, future work will focus on integrating anomaly detection algorithms to ensure the reliability of the design space~\cite{benavides2010automated}. Future research could focus on refining graph structure models and addressing limitations in current methodologies to improve their application in various domains \textcolor{black}{or adding user interface to define the graph with visual tools.}
For instance, the role graph in this work is designed as a DAG to prevent cyclical dependencies, ensuring that meta-decreed variables do not influence one another. While DAGs provide clear hierarchical structures and efficient traversal~\cite{bender2005lowest}, they limit the representation of complex interdependencies. The DSG addresses this by recursively parsing design variables in a flexible, non-linear sequence~\cite{bussemaker2024ADSG}. Future research will focus on expanding the DSG to incorporate more complex structures, including cyclic dependencies, leveraging advancements like hypergraphs~\cite{angles2018introduction} and cyclic graph models~\cite{bang2009disjoint} to enhance the framework's applicability to sophisticated optimization and surrogate modeling challenges. However, enumerating all valid instances within such complex structures quickly becomes computationally intractable. To overcome this, advanced operations research techniques will be required to efficiently estimate the number of possibilities and explore these graphs effectively~\cite{papadimitriou1998combinatorial}. Another promising direction for future development is the generalization of the framework to enable more automatic learning. For instance, relationships between variables could be learned directly from data~\cite{bourdais2024codiscovering}. Moreover, adopting more general hierarchical representations could help tackle problems involving partial and incomplete data, particularly when numerous variables exist but are not fully observed~\cite{owhadi2022computational}.
Also, we can extend relationship definitions to include any function of the form $R(\phi(x_1), \psi(\textcolor{black}{x_2}$)).

A further area of interest lies in reducing graph complexity to improve scalability. Future work will explore techniques such as kernelization of the vertex cover problem~\cite{Buss} and propositionalization methods, which convert graph structures into flat propositions, making it easier to apply distance metrics over propositional data~\cite{karunaratne2009graph}. These strategies will enhance the framework’s scalability when dealing with large, complex design spaces. Finally, the framework will be adapted to tackle highly dimensional problems with architectural constraints. This will involve using advanced techniques such as KPLS combined with hierarchical kernels~\cite{Mixed_Paul_PLS,Bouhlel18}. The goal is to enable the framework to handle more complex optimization tasks by integrating dimension reduction techniques with structural flexibility. 

\textcolor{black}{As future work, we plan to integrate the Design Space Graph within coupled system XDSM, thereby extending graph-structured design space for the treatment of coupling variables by explicitly tracking their relationships across surrogate-based co-simulation chains~\cite{dubreuil2021development}. Still, our work is general and can be leveraged for many applications beyond coupled system. For instance, the distance introduced in this work is model-agnostic and applicable to a wide range of surrogate models beyond Gaussian processes~\cite{halle2024graph}.
Specifically, it can be used as a dissimilarity measure for instance-based learners, to construct alternative correlation kernels, or to generate finite-dimensional embeddings that produce feature vectors suitable for any parametric regressor. Future work will extend this framework to higher-dimensional problems and explore its integration with $k$-nearest neighbors, neural networks, and inverse-distance weighting surrogate models.
Moreover, our methodology aligns with the principles of systems engineering by providing a general structured modeling procedure that spans from architecture definition to surrogate construction~\cite{chan2023goal}.}

\textcolor{black}{This framework formalizes the translation of component- and function-level dependencies into graph-structured design spaces, thereby reducing ambiguity and enabling analytical optimization. Consequently, future research will investigate the integration of this approach within system architecture design and optimization reasoning environments~\cite{bussemaker2022adore} and its application to co-design and socio-technical simulation problems~\cite{saves2025surrogate}. Beyond formal representation, the proposed design space graphs and hierarchical distances directly support surrogate modeling and optimization by providing consistent variable embeddings and correlation structures. The framework can also interface with original simulations whenever hierarchical dependencies must be explicitly managed in Design of Experiments or system analysis workflows.
This work is general and can be applied to any kind of complex system that involves a structured design space or heterogeneous datasets. }

\section*{Replication of results}
This framework is implemented in the open-source SMT toolbox~\cite{saves2023smt}, extending its capabilities to support hierarchical surrogate models. This implementation is documented in a dedicated tutorial notebook {\url{https://github.com/SMTorg/smt-design-space-ext/blob/main/tutorial/SMT_DesignSpace_example.ipynb}}, easing its practical application. Also one can automatically define a DSG from the variables specifications {\url{https://github.com/SMTorg/smt-design-space-ext/}}.

\section*{Fundings}
\noindent  This research is funded by a Natural Sciences and Engineering Research Council of Canada (NSERC) PhD Excellence Scholarship (PGS D), a Fonds de Recherche du Qu\'ebec (FRQNT) PhD Excellence Scholarship and an Institut de l'{\'E}nergie Trottier (IET) PhD Excellence Scholarship, 
as well as by the NSERC discovery grant  RGPIN-2024-05093.
This work is part of the activities of ONERA - ISAE - ENAC joint research group. The research presented in this paper has been performed in the framework of the COLOSSUS project (Collaborative System of Systems Exploration of Aviation Products, Services and Business Models) and has received funding from the European Union Horizon Europe program under grant agreement n${^\circ}$ 101097120 and in the MIMICO research project funded by the Agence Nationale de la Recherche (ANR) n$^o$ ANR-24-CE23-0380. 

\section*{Conflict of interest statement}
The author declares no Conflict of interest.

\section*{Ethics approval and Consent to participate}  
Not applicable.

\section*{Data Availability}
All source code and demonstration data are openly available at {\url{https://github.com/SMTorg/smt-design-space-ext/}}.  Additional datasets are available on request.

\section*{Author Contributions}
P. Saves: conceptualization, software, writing, formal modeling;  
E. Hallé‑Hannan: experimental analysis, writing, outline;
J. Bussemaker: formal modeling, Contributions drafting;  
Y. Diouane: supervision, review;  
N. Bartoli: supervision, review.  
All authors approved the final manuscript.

\appendix

\section{Data representation of structured spaces}
\label{sec:data_representation}

In general, structured representations can be categorized into three major types: graph-based, logic-based, and frame-based representations~\cite{ontanon2020overview}, although some models fall outside this classification (\textit{e.g.}, general matrices~\cite{GrosseMatrix}).

Graph-based representations are particularly powerful in fields such as structural pattern recognition and network analysis. In these models, data is represented as graphs, where nodes correspond to entities and edges to the relationships between them. This approach is highly valued for its invariance to transformations, making it robust in various applications where the structure of the data is as important as the content itself~\cite{gao2010survey}.  
Recent advances in causal representation learning extend these models by approximating latent causal structures from high-dimensional data~\cite{bourdais2024codiscovering}. This data-driven approach moves beyond correlations and supports out-of-distribution generalization and planning~\cite{von2024nonparametric,Payandeh2023137621}.  

Logic-based representations are another form of structured representation, widely used in domains such as inductive logic programming and semantic web technologies. These models use logical statements to represent data, enabling complex inferences and generalizations~\cite{mitchell1986explanation}. 
The key advantage of logic-based representations lies in their ability to encode generalization and inference naturally, allowing for the incorporation of background knowledge through rules. This makes them particularly effective in domains requiring complex reasoning and knowledge representation. For instance, within a data landscape, logic-based representations can show semantic relationships and logical dependencies across different systems or domains, enhancing the ability to spot overarching patterns in knowledge~\cite{gonzalez2021case}.
Horn clauses, description logics, and order-sorted feature structures are key logic-based formalisms. Horn clauses enable software verification and automated theorem proving~\cite{de2022analysis}. Description logics balance expressive power and computational efficiency, sitting between propositional and first-order logic~\cite{DLbook}. Order-sorted feature structures, a subset of first-order logic, are used in case-based reasoning and natural language processing~\cite{kaneiwa2004order}. 

Frame-based representation is a classical approach to knowledge modeling, commonly used in domains like case-based reasoning and Artificial intelligence~\cite{Minsky}. Knowledge is organized using attribute-value pairs within a hierarchical structure, where each frame represents a concept or entity. This structure allows properties to be inherited from parent frames, facilitating efficient representation of complex knowledge~\cite{song2022knowledge}. Frames also support default values and data types, enhancing their utility in knowledge-based systems~\cite{song2022knowledge}.
Frames enable reasoning and inference, making them valuable for tasks like automated theorem proving and decision-making~\cite{Minsky}. Though similar to other methods like knowledge graphs~\cite{kejriwal2022knowledge}, semantic networks~\cite{quillian1969teachable}, and concept maps~\cite{novak1990concept}, frame-based systems remain integral to structured knowledge representation, particularly in case-based reasoning and statistical relational learning~\cite{gonzalez2021case}.

Structured data representations play a crucial role across various domains of machine learning and artificial intelligence, each tailored to the specific demands of different tasks and applications. Recent advances in supervised learning of representations have shown great promise in automatically extracting task-specific features from structured spaces, especially in categorical and graph-based data. These methods not only enhance flexibility but also significantly improve the predictive performance of models~\cite{Wu20161424,Oh202438541}. Moreover, supervised learning approaches have been particularly effective in identifying meaningful, task-driven features from complex data structures, offering robust solutions in fields like graph-based modeling and few-shot learning~\cite{Esser202411910,Chen20237069}. As research progresses, selecting the appropriate data representation remains a critical factor, as it directly influences the effectiveness of learning algorithms and their ability to generalize across various applications. For example, deep representation learning continues to expand in its ability to handle structured data, which poses unique challenges due to its high dimensionality and inherent complexity~\cite{Payandeh2023137621}.

\textcolor{black}{A critical data-issue when working with structured spaces is the presence of \textit{missing values} and the mechanism behind them. Following Rubin’s classical taxonomy~\cite{rubin1976inference}, missingness is characterized as Missing Completely At Random (MCAR) when the fact of missing does not depend on any observed or unobserved data, Missing At Random (MAR) when the missingness depends only on observed values, and Missing Not At Random (MNAR) when missingness depends on the unobserved (missing) values themselves which is exactly the case in our study~\cite{bussemaker2024system}.
In many high-dimensional or structured-data settings (\textit{e.g.}, graphs, feature matrices, heterogeneous data types), these distinctions remain critical for modeling, imputation, and inference, because the mechanism determines whether standard approaches (\textit{e.g.}, deletion, mean‐imputation, matrix‐completion) are valid or biased. In particular, in our case, under MNAR, one must explicitly model the missingness mechanism or risk biased learning or poor generalization~\cite{josse2012handling}.
Recognizing and properly handling missingness is thus important in representation learning from structured spaces, for example, when elements of a graph (nodes, edges) or frame attributes are missing non-randomly, this can distort learned embeddings or downstream predictions~\cite{josse2024consistency}. A major interest of such statistical tools in our context is that they can be used if the graph structure is both implicit and unknown to uncover it without any knowledge or specifications.}


\section{SMT Design Space Graph implementation usage}
\label{app:pseudo-code}
\begin{figure}[H]
\hspace{0.01cm}
\scalebox{0.95}{
    \includegraphics[width=1\textwidth]{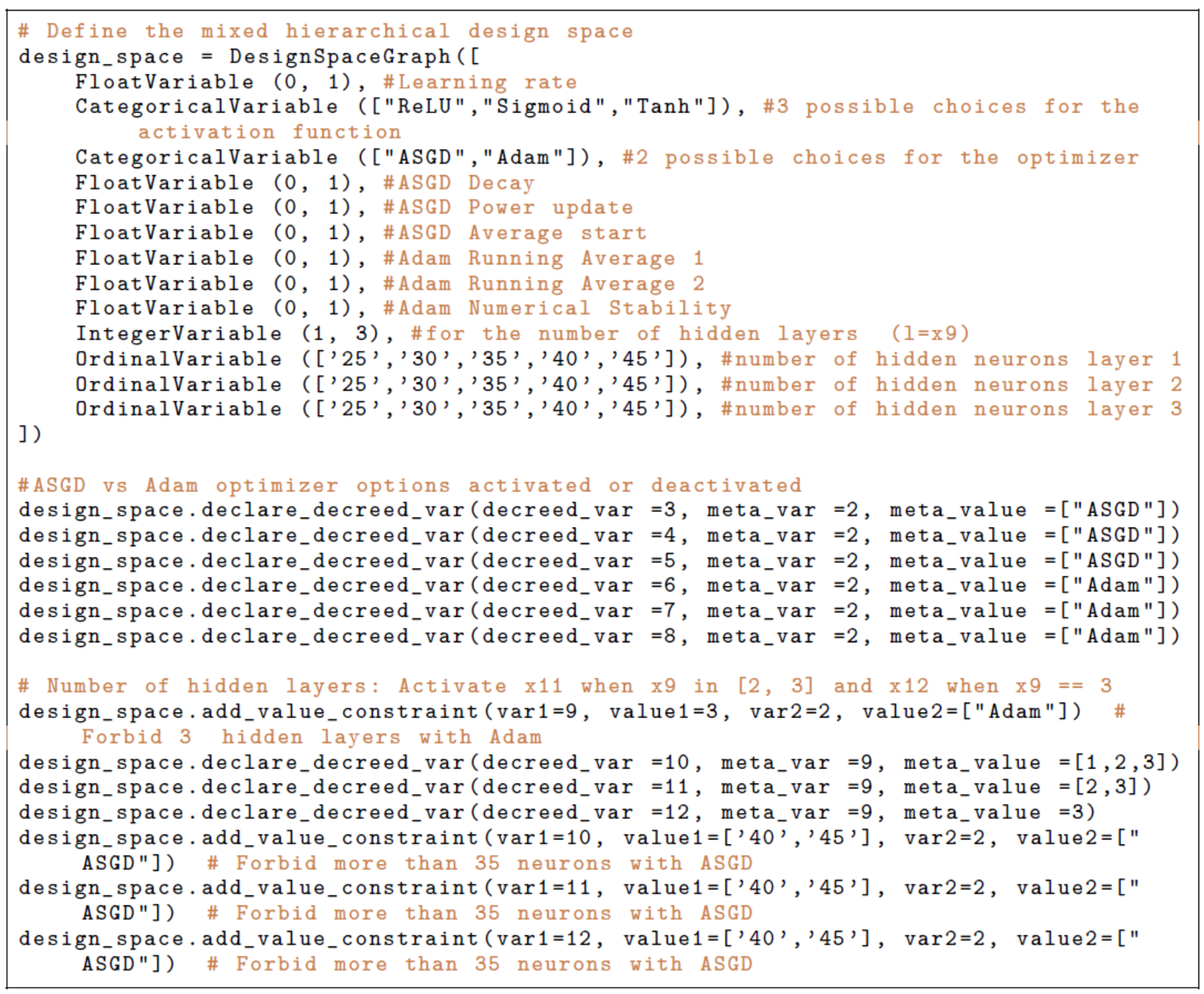}
}
\caption{\texttt{AdsgDesignSpaceImpl} definition for the neural network problem.}
\label{fig:ds_nn}
\end{figure}

In this section, we present, in~\figref{fig:ds_nn}, the code that generates the hierarchical design space graph of the Multi-Layer Perceptron hierarchical design space. This design space is adapted for visualization, distance computation, and surrogate modeling. \textcolor{black}{Recall that the best option for architecture modeling is the DSG of the adsg-core toolbox that is fast and gives explicit visualizations of the DSG in DOT language, draw.io, or GML~\cite{ellson2002graphviz}, and our main software contribution is to allow for combining DSG with SMT to define hierarchical surrogate models.
}
\textcolor{black}{For architecture modeling, we recommend the DSG implementation provided by the \texttt{adsg-core} library: it extends the DSG concept with model-based systems engineering primitives. Also, this implementation offers a well-documented library API for programmatic graph construction and can export graphs into DOT language, draw.io, and GML formats~\cite{ellson2002graphviz}. 
Moreover, ADORE is a user-friendly GUI for building, visualizing, and wiring ADSG models to evaluation and optimization backends; however, exhaustive enumeration of very large configuration families remains an implementation and performance challenge governed by the ADSG APIs and the chosen encoder implementations. 
\textcolor{black}{More details on the software, modeling, and model-based system engineering are given in~\cite{bussemaker2025system}.}
\textcolor{black}{Note that in the current implementation, the DSG encoder/decoder used within the ADSG framework assumes a directed acyclic graph and does not attempt to resolve cyclic dependencies. Consequently, cyclic or implicit relationships between variables are not supported by the present system and are outside the scope of this paper. Exhaustive enumeration of very large configuration families, as well as the detection and resolution of cyclic dependencies, remain implementation and performance challenges.}
Still, in practice, the ADORE/ADSG/DSG toolchain automates generation and sampling of parameterized families, while the encoder choice (complete vs. fast/lazy) controls whether combinations are enumerated exhaustively or handled through more scalable decoding and correction strategies~\cite{bussemaker2024ADSG,bussemaker2025system}.}

\lstset{
language=Python,
morekeywords={smt}
emphstyle=\color{red}, 
basicstyle=\ttfamily \footnotesize,
frame=single,
showspaces=false,
showstringspaces=false,
keywordstyle=\ttfamily \color{black},
commentstyle=\color{green!30!red!70}\ttfamily,
breaklines=true,
}

\lstset{
language=Python,
morekeywords={smt}
emphstyle=\color{red}, 
basicstyle=\ttfamily \small,
frame=single,
showspaces=false,
showstringspaces=false,
keywordstyle=\ttfamily \color{black},
commentstyle=\color{green!30!red!70}\ttfamily,
breaklines=true,
}

\section{Symmetric Positive Definiteness of hierarchical kernels}
\label{app:SPD}
In this section, we show that our proposed hierarchical kernel—built from neutral, meta‐level, and decreed‐variable components—yields a symmetric positive‐definite covariance function suitable for Gaussian‐process models on arbitrarily nested, mixed‐type variable spaces.
The new kernel $k$ that we propose for hierarchical variables extends a previous work~\cite{saves2023smt} and is given by
\begin{eqnarray} 
k(\bold{X}, \bold{X'}) =& k_{\neutral} (\bold{X}_{\neutral},\bold{X'}_{\neutral}) \times  
k_\meta(\bold{X}_\meta,\bold{X'}_\meta) \times  \nonumber\\
& \ k_{\meta,\decreed} ( [\bold{X}_\meta, \bold{X}_{\acting}(\bold{X}_\meta)], [\bold{X'}_\meta, \bold{X'}_{\acting}(\bold{X'}_\meta)]), 
\label{eq:hier_ker}
\end{eqnarray}
where $k_{\neutral}$, $k_\meta$ and $k_{\meta,\decreed}$ are as follows:
 \begin{itemize}
 \setlength\itemsep{0cm}
\item  $k_{\neutral}$ represents the neutral kernel that encompasses both categorical and quantitative neutral variables, \textit{i.e.}, $k_{\neutral}$ can be decomposed into two parts $k_{\neutral} (\bold{X}_{\neutral},\bold{X'}_{\neutral})= k^{\cat}(\bold{X}_{\neutral}^{\cat},\bold{X'}_{\neutral}^{\cat}) \times  k^{\quant} (\bold{X}_{\neutral}^{\quant},\bold{X'}_{\neutral}^{\quant}).$
The categorical kernel, denoted $k^{\cat}$, could be any Symmetric Positive Definite (SPD)~\cite{Mixed_Paul} mixed kernel. 
For the quantitative (integer or continuous) variables, a distance-based kernel is used. 
The chosen quantitative kernel (\textit{e.g.}, exponential or Matérn) always depends on a given distance $d$.
For example, the $n$-dimensional exponential kernel gives 

\begin{equation}  
 k^{\quant}(\bold{X}^{\quant},\bold{X'}^{\quant}) = \displaystyle{ \prod^{n}_{i=1} \exp ( - d(\bold{X}^{\quant}_i,\bold{X'}^{\quant}_i))}.
\end{equation}
    \item $k_{\meta}$ is the meta variables related kernel.
    It is also separated into two parts:
$k_{\meta} (\bold{X}_{\meta},\bold{X'}_{\meta})= k^{\cat}(\bold{X}_{\meta}^{\cat},\bold{X'}_{\meta}^{\cat})k^{\quant} (\bold{X}_{\meta}^{\quant},\bold{X'}_{\meta}^{\quant})$ where the quantitative kernel is ordered and not continuous because meta variables take value in a finite set.  
\item  $k_{\meta,\decreed}$ is an SPD kernel that models the correlations between the meta levels (all the possible subspaces) and the decreed variables. It is a quantitative kernel depending on the aforementioned graph-structure distance $ \edist$. In particular, we have developed several continuous relaxations to make any distance, such as the new graph-structure one, to induce a categorical kernel. This allows us to generalize here the previous kernel~\cite{saves24} to also handle decreed categorical variables as described in~\cite{bussemaker2024system}.
\end{itemize}

\begin{theorem}
The kernel $K^{naive}$ defined as follows is not a SPD kernel. 
\begin{equation}
 \begin{cases}
  K^{naive}(\bold{X},\bold{X'}) = k_{\decreed}(\bold{X}_{\acting} (\bold{X'}_\meta), \bold{X'}_{\acting} (\bold{X'}_\meta)) k_{\neutral}(\bold{X}_{\neutral}, \bold{X'}_{\neutral}) , & \text{if}\ \bold{X}_\meta = \bold{X'}_\meta \\  
  K^{naive}(\bold{X},\bold{X'}) =  k_{\meta}(\bold{X}_\meta, \bold{X'}_\meta) k_{\neutral}(\bold{X}_{\neutral}, \bold{X'}_{\neutral}), & \text{if}\ \bold{X}_\meta \neq \bold{X'}_\meta \\
\end{cases} 
\end{equation}
\end{theorem}

\begin{proof}

$K^{naive}$ is a kernel if and only if $K^{naive}$ can be written as $K^{naive} (\bold{X},\bold{X'}) = \kappa ( d(\bold{X},\bold{X'}) ) $ with $ \kappa $ being an SPD function and $d$ being a distance in a given Hilbert space.
If the two continuous decreed inputs $\bold{X}_{dec}$ and $\bold{X'}_{dec}$ are in the same subspace ($\bold{X}_\meta=\bold{X'}_\meta$), then  $k_\meta(\bold{X}_\meta,\bold{X}_\meta) =1$ and $ K^{naive}(\bold{X},\bold{X'}) = k_{\decreed}(\bold{X}_\acting(\bold{X}_\meta), \bold{X'}_\acting(\bold{X'}_\meta) )$. This works without any problem, a categorical variable being fully correlated with itself. 

On the contrary, if the two continuous inputs $\bold{X}_{\decreed}$ and $\bold{X'}_{\decreed}$ lie in two different subspaces ($\bold{X}_\meta \neq \bold{X'}_\meta$), then, we have $ k_{dec}( \bold{X}_{\decreed}, {\bold{X'}}_{\decreed} ) =1 \implies  d(\bold{X}_{\acting} (\bold{X}_\meta), \bold{X'}_{\acting} (\bold{X'}_\meta) ) = 0 $ .

We know that a distance is defined such that $ d(\bold{X},\bold{X'}) = 0 \Leftrightarrow  \bold{X} = \bold{X'}  $ (Identity of indiscernibles).
Yet, we have  $ d(  \bold{X}_{\acting} ( \bold{X}_\meta),  \bold{X'}_{\acting} ( \bold{X'}_\meta) ) = 0 $ and  $ \bold{X}_{\acting} ( \bold{X}_\meta) \neq   \bold{X'}_{\acting}( \bold{X'}_\meta)$. $d$ is not a distance because it is always equal to $0$ for points in distinct decreed spaces. Therefore, the kernel vanishes to 1, and the correlation matrix is degenerated and cannot be SPD.

Based on this proof, the idea of the arc-kernel is to have a constant residual distance between distinct subspaces. In other words, there is a non-zero distance between $\mathcal{X}( \bold{X}_\meta)$ and $\mathcal{X}( \bold{X'}_\meta)$ if $ \bold{X}_\meta \neq  \bold{X'}_\meta  $.

\end{proof}

\begin{theorem}
Our \texttt{Alg-Kernel} kernel is SPD and so is $k$ defined in~\eqnref{eq:hier_ker}.\end{theorem}
\begin{proof}

Let $I_{\bold{X}}$ be the subset of indices $ i \in I_\decreed$ that are decreed-included by $  \bold{X}_{\meta}$ such that $ I^{inter}_{\bold{X},\bold{X'}}  = I_{\bold{X}} \bigcap I_{\bold{X'}}$. Let $d_E ( [x_\bold{X},x_\bold{X'}],[y_\bold{X},y_\bold{X'}] ) = \theta_i \sqrt{ ( x_\bold{X} -x_\bold{X'} )^2 +(y_\bold{X}- y_\bold{X'} )^2 }  $ be an Euclidean distance in $\mathbb{R}^2  \times \mathbb{R}^2  $.
Due to~\cite[Proposition 2]{Hutter}, we only need to show that, for any two inputs $\bold{X},\bold{X'} \in \mathcal{X}$, the isometry condition $d_E (f^{alg}_{i} (\bold{X}) ,f^{alg}_{i} (\bold{X'} )) $  holds for a given function $f^{alg}_{i}$, that is equivalent to having a Hilbertian metric. In other words, $d_E$ is isomorphic to an $L^2$ norm~\cite{haasdonk2004learning}. Such kernels are well-known and referred to as "substitution kernels with Euclidean distance"~\cite{sow2023learning}.

$\forall i \in I_\decreed, f_i^{alg}( \bold{X}_{\decreed}) $ is defined by
\begin{equation}
 \begin{cases}
      f_i^{alg}( \bold{X}_\acting ) = [\frac{1-((   \bold{X}_{dec})_i)^2}{1+((   \bold{X}_{dec})_i)^2 },\frac{2 (   \bold{X}_{dec})_i   }{1+ (( \bold{X}_{dec})_i)^2 }]  \text{ if } i \in I_{u}  \\
      f_i^{alg}( \bold{X}_{\decreed}) = [0,0] \text{ otherwise }  \\
\end{cases} 
\end{equation}  

Case 1: $i \in I_\decreed, i \notin I_{\bold{X}}, i \notin I_{\bold{X'}} $. 
$$d_E ( f_i^{alg} ( \overline{\bold{X}}_{dec} ) ,f_i^{alg}( \overline{\bold{X'}}_{dec}  )) =d_E ([0,0],[0,0]) = 0. $$
Therefore, the complementary space is not relevant as for the arc-kernel. \medbreak

Case 2: $i \in I_\decreed, i \in I_{\bold{X}}, i \notin I_{\bold{X'}} $.

$$d_E ( f_i^{alg}(  \bold{X}_{dec} ) ,f_i^{alg} (  \overline{\bold{X'}}_{dec})) =d_E ( [\frac{1-((   \bold{X}_{dec})_i)^2}{1+((   \bold{X}_{dec})_i)^2 },\frac{2 (   \bold{X}_{dec})_i }{1+ (( \bold{X}_{dec})_i)^2 }] , [0,0] ) $$
$$=  \theta_i . $$
This case corresponds to the kernel $K_\meta^{alg}( \bold{X}_\meta, \bold{X'}_\meta)$ when $ \bold{X}_\meta \neq  \bold{X'}_\meta $.  \medbreak

Case 3:  $i \in I_\decreed, i \in I^{inter}_{\bold{X},\bold{X'}}$.

\begin{equation*}
    \begin{split}
         &d_E ( f_i^{alg}(  \bold{X}_{dec}  ) ,f_i^{alg}(  \bold{X'}_{dec}  ))  \\
        &= d_E \left( \left[  \frac{1-((   \bold{X}_{dec})_i)^2}{1+((   \bold{X}_{dec})_i)^2 },\frac{2 (   \bold{X}_{dec})_i }{1+ (( \bold{X}_{dec})_i)^2 } \right], \left[\frac{1-((   \bold{X'}_{dec})_i)^2}{1+((   \bold{X'}_{dec})_i)^2 },\frac{2 (   \bold{X'}_{dec})_i }{1+ (( \bold{X'}_{dec})_i)^2 }\right] \right) \\
        &= 2 \theta_i \frac{ |( \bold{X}_{dec})_i- ( \bold{X'}_{dec})_i|}{ \sqrt{{( ( \bold{X}_{dec})_i })^2+1}\sqrt{{ (( \bold{X'}_{dec})_i })^2+1}}
    \end{split}
\end{equation*}
This case corresponds to the kernel $K_{dec}^{alg}( \bold{X}_{dec}( \bold{X}_m), \bold{X'}_{dec}( \bold{X'}_m))$ when $ \bold{X}_m =  \bold{X'}_m $.  \medbreak

Therefore, there exists an isometry between our algebraic distance and the Euclidean distance. The distance being well-defined for any given kernel, the matrix obtained with our model is SPD. 

Although in~\cite{Hutter}, they also demonstrate the metric property of their distance formally. The non-negativity and symmetry of $d^\text{alg}$ are trivially proven knowing that the hyperparameters $\theta$ are strictly positive. To prove the triangle inequality, consider $(u,v,w) \in \mathcal{X}^3.$

Case 1: $i \in I_\decreed, i \notin I_{u}, i \notin I_{v} $. 
$$d^\text{alg} (u_i,v_i) = 0 \leq d^\text{alg} (u_i,w_i) + d^\text{alg} (w_i,v_i) \text{ by non negativity.}$$   \medbreak

Case 2: $i \in I_\decreed, i \in I_{u}, i \notin I_{v} $.

\begin{itemize} [leftmargin=3cm]
    \item $   i \in I_{w}, \ d^\text{alg} (u_i,v_i) = \theta_i \leq d^\text{alg} (u_i,w_i)  + \theta_i\text{ by non negativity.}$   
 \item $  i \notin I_{w}, \ d^\text{alg} (u_i,v_i) = \theta_i \leq \theta_i  + d^\text{alg} (w_i,v_i)  \text{ by non negativity.}$   

\end{itemize}
\medbreak

Case 3: $i \in I_\decreed, i \in I^{inter}_{u,v}$.

\begin{itemize} [leftmargin=3cm]
    \item $  i \notin I_{w}, \ d^\text{alg} (u_i,v_i) = 2 \theta_i \frac{ |u_i- v_i|}{ \sqrt{{( u_i })^2+1}\sqrt{{ (v_i })^2+1}}  \leq  \theta_i  + \theta_i \ \text{ for } (u_i,v_i) \in [0,1]^2.$   
 \item $  i \in I_{w}, \text{ Knowing that}  \frac{ |a- c|}{ \sqrt{{ a }^2+1}\sqrt{{ c }^2+1}} +\frac{ |c- b|}{ \sqrt{{ c }^2+1}\sqrt{{ b}^2+1}}   \geq \frac{ |a- b|}{ \sqrt{{ a }^2+1}\sqrt{{ b }^2+1}}   $, we have\\ $\ d^\text{alg} (u_i,v_i) =  2 \theta_i \frac{ |u_i- v_i|}{ \sqrt{{( u_i })^2+1}\sqrt{{ (v_i })^2+1}}   \leq  d^\text{alg} (u_i,w_i) +  d^\text{alg} (w_i,v_i) $.
\end{itemize}

Using~\cite[Theorem 3]{halle2024graph}, any graph-structure distance is well-defined if and only if $\sup \{ d \left(\mu, \nu \right): \mu, \nu \in  \mathcal{X}^2 \}/2 \leq \theta_i $. This can make the proof easier and faster as follows. 
First, we can compute the maximum of the algebraic distances which gives $\text{max}_{(x,y) \in \mathbb{R}^2} \ d^{\text{alg}} (x,y) = \text{max}_{(x,y) \in \mathbb{R}^2}  \ 2 \theta_i |x-y| \sqrt{\frac{1}{x^2+1}} \sqrt{\frac{1}{y^2+1}} = d^{\text{alg}}(x,-\frac{1}{x}) = 2 \theta_i$. Then, we just have to check that $ 2 \theta_i /2 \leq \theta_i$, which is the case. This theorem brings two interesting insights on our distance, first, it is a limit case as the distance between two non-interacting subspaces is properly maximized, and second, the algebraic circle-based distance is well-constructed as the maximal distance corresponds to the maximal arc of 1 radian, making it a generalization of the Arc-distance~\cite{Zaefferer}.


\medbreak

\end{proof}





\pdfbookmark[1]{References}{sec-refs}

\end{document}